\documentclass[letterpaper,journal]{IEEEtran}

\usepackage{amssymb,amsmath,amsfonts}
\usepackage{algorithm}
\usepackage{algpseudocode}
\usepackage{array}
\usepackage{bbold}
\usepackage{booktabs}
\usepackage{colortbl}
\usepackage[hidelinks]{hyperref}
\usepackage{graphicx}
\usepackage{makecell}
\usepackage[mathlines, switch]{lineno}
\usepackage{mathtools}
\usepackage{multirow}
\usepackage[sort&compress, numbers]{natbib}
\usepackage{pifont}
\usepackage{siunitx}
\usepackage{stfloats}
\usepackage[caption=false,font=footnotesize]{subfig}
\usepackage{textcomp}
\usepackage{xurl}
\usepackage{verbatim}
\usepackage[dvipsnames]{xcolor}

\newcommand{\cmark}{\ding{51}}
\newcommand{\xmark}{\ding{55}}
\newcommand{\pmark}{$\triangle$}

\hypersetup{
  colorlinks   = true, 
  urlcolor     = CadetBlue, 
  linkcolor    = Cerulean, 
  citecolor   = Maroon 
}

\begin{document}

\title{RoadWeaver: Large-Scale Lane-Level HD Map Generation from Scratch for Autonomous Driving Simulation}

\author{Yueyuan~Li, Zexi~Chen, Weijie~Xi, Mingyang~Jiang, Songan~Zhang, Hanyang~Zhuang, and~Ming~Yang 
}

\maketitle


\begin{abstract}

Autonomous driving simulation requires diverse and scalable lane-level HD maps to support long-horizon evaluation across complex road networks.
Existing approaches either rely on handcrafted or reconstructed real-world maps, which limits scalability, or generate only local road structures rather than complete HD maps.
We present RoadWeaver, a coarse-to-fine framework for from-scratch generation of diverse, large-scale HD maps.
RoadWeaver first synthesizes a global road layout, expands it into a connected road network, and then constructs lane-level geometry with topologically consistent lane connectivity.
Experimental results show that RoadWeaver achieves a 99.8\% reachability, a 10.7\% dead-end ratio, and an endpoint alignment error of 0.24 m.
Compared with SOTA generation methods, it reduces endpoint alignment error by 94.4\% while generating complete HD maps in 1.39--3.50 s.
The generated maps can be directly deployed in driving simulators, providing scalable simulation environments for future closed-loop evaluation of autonomous driving systems.
The training code and an out-of-the-box implementation of RoadWeaver will be released upon acceptance.

\end{abstract}



\section{Introduction}\label{section:introduction}

Simulation has become an indispensable tool for the development and evaluation of autonomous driving systems (ADS)~\cite{li2024choose}.
Widely adopted benchmarks, such as the CARLA Leaderboard and Bench2Drive, have highlighted the importance of evaluating ADS over long-horizon closed-loop routes rather than isolated scenario fragments~\cite{dosovitskiy2017carla, jia2024bench2drive}.
Such evaluations expose planning failures that accumulate over time, including inappropriate lane selections, route deviations, and traffic-rule violations that may remain hidden in short, independent scenarios. 
Consequently, long-horizon evaluation demands road maps that are not only large-scale and geographically diverse, but also sufficiently detailed to accurately represent lane-level geometry and connectivity.

\begin{figure}[tb]
    \centering
    \includegraphics[width=\linewidth]{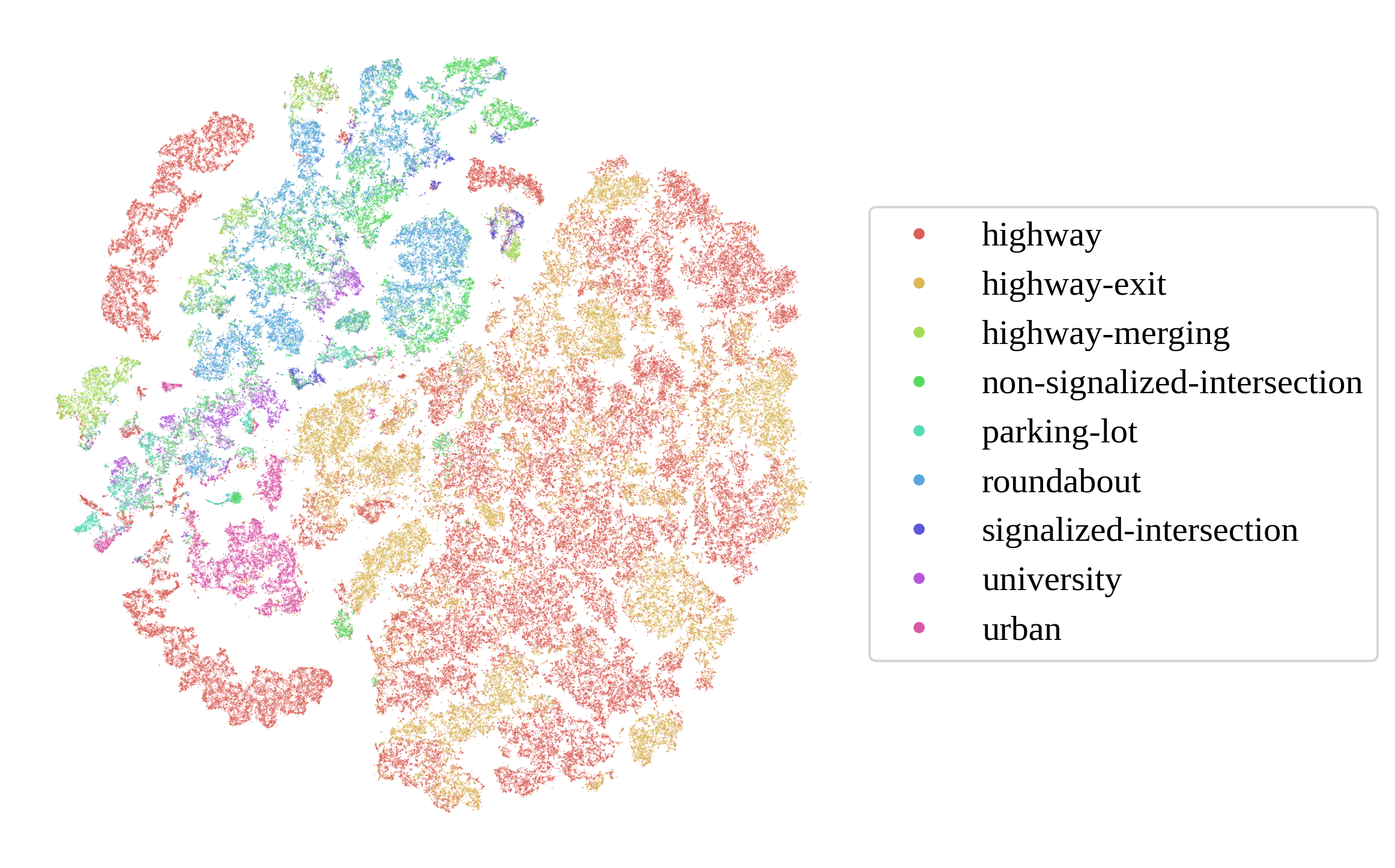}
    \caption{t-SNE visualization of driving behavior analysis collected from multiple autonomous driving datasets, colored by scenario type.}\label{figure:t-sne}
\end{figure}

Unlike conventional road networks designed primarily for traffic-flow analysis, high-definition (HD) maps provide rich geometric, semantic, and topological information that is essential for ADS evaluation.
Among these properties, road-network topology plays a crucial role in shaping driving behavior.
To better understand this relationship, we analyze trajectory dynamics, including velocity, acceleration, steering, and other motion characteristics, from multiple autonomous driving datasets~\cite{coifman2017critical,krajewski2018highd,bock2020ind,krajewski2020round,bock2021highly,zhan2019interaction,caesar2021nuplan,ettinger2021large,shen2022parkpredict,zheng2024citysim}. 
As illustrated in Fig.~\ref{figure:t-sne}, trajectories from different road topologies exhibit distinct behavioral distributions, indicating that topology influences driving and interaction patterns. 
These observations suggest that increasing the diversity of road-network topology is crucial for exposing a broader spectrum of driving behaviors and improving the structural coverage of long-horizon ADS evaluation.

Existing simulation maps are primarily obtained through three approaches: handcraft~\cite{dosovitskiy2017carla,highway-env}, reconstructed from real-world~\cite{caesar2021nuplan,ettinger2021large}, and automatic generation~\cite{li2022metadrive,mi2021hdmapgen}. 
The first two approaches suffer from limited scalability: manually designed maps require substantial engineering effort, whereas reconstructed HD maps are restricted to existing geographic regions and their inherent topological characteristics. 
Automatic map generation offers a scalable alternative for constructing large-scale driving environments.
However, existing methods still struggle to simultaneously achieve global topological coherence, lane-level geometric validity, controllable road morphology, and direct compatibility with mainstream simulation platforms. 
These limitations make it difficult to generate diverse road networks suitable for systematic long-horizon ADS evaluation.

To address these challenges, we propose RoadWeaver, a framework for generating diverse, large-scale lane-level HD maps from scratch.
RoadWeaver constructs globally consistent road networks while preserving lane-level geometric correctness and enabling flexible control over road morphology.
The generated maps can be exported directly to OpenStreetMap (OSM) and OpenDRIVE formats~\cite{haklay2008openstreetmap,dupuis2010opendrive}, allowing seamless integration with existing autonomous driving simulators.
We further integrate RoadWeaver into Tactics2D~\cite{li2024tactics2d}, providing an out-of-the-box map generation pipeline for scalable simulation and customized long-horizon ADS evaluation.

The contributions of this work are summarized as follows:

\begin{itemize}

\item We propose RoadWeaver, a from-scratch framework for generating diverse lane-level HD maps with valid road-network topology and lane geometry.

\item RoadWeaver supports large-scale map generation by controlling the spatial extent while maintaining a target road-network density.

\item RoadWeaver is simulation-ready, exporting maps directly to OSM and OpenDRIVE formats and integrating seamlessly with Tactics2D for closed-loop ADS evaluation.

\end{itemize}


\section{Related Works}\label{section:related-works}

Existing approaches to preparing maps for autonomous driving simulation can be broadly categorized into three groups: manual construction, reconstruction from real-world observations, and automatic generation.
The last category can be further divided into rule-based and data-driven methods.
Table~\ref{table:related-works} summarizes the capabilities of representative methods.


\begin{table*}[tb]
\centering
\caption{Comparison of representative methods for autonomous driving map preparation.}\label{table:related-works}
\begin{tabular}{l|cc|c|ccc}
\toprule
\multirow{2}{*}{\textbf{Method}} 
& \multicolumn{2}{|c|}{\textbf{Diversity}} 
& \textbf{Scale}
& \multicolumn{3}{c}{\textbf{Usability}} \\

& \textbf{From Scratch}
& \textbf{Global Topology}
& \textbf{Large-scale}
& \textbf{Lane-level}
& \textbf{Controllable}
& \textbf{Simulation-ready} \\
\midrule

CARLA (2017)~\cite{dosovitskiy2017carla}
& \xmark & \cmark & \xmark & \cmark & \xmark & \cmark \\

Highway-env (2020)~\cite{highway-env}
& \xmark & \pmark & \xmark & \pmark & \pmark & \cmark \\

HDMapGen (2021)~\cite{mi2021hdmapgen}
& \cmark & \pmark & \xmark & \cmark & \xmark & \xmark \\

HDMapNet (2022)~\cite{li2022hdmapnet}
& \xmark & \xmark & \xmark & \cmark & \xmark & \xmark \\

MapTR (2022)~\cite{liao2022maptr}
& \xmark & \xmark & \xmark & \cmark & \xmark & \xmark \\

VectorMapNet (2023)~\cite{liu2023vectormapnet}
& \xmark & \xmark & \xmark & \cmark & \xmark & \xmark \\

MetaDrive (2022)~\cite{li2022metadrive}
& \cmark & \cmark & \cmark & \pmark & \cmark & \cmark \\

RoadGen (2024)~\cite{yang2024roadgen}
& \cmark & \cmark & \pmark & \pmark & \cmark & \pmark \\

ControlMap (2026)~\cite{farag2026controlmap}
& \cmark & \pmark & \xmark & \cmark & \cmark & \xmark \\

\midrule
RoadWeaver
& \cmark & \cmark & \cmark & \cmark & \cmark & \cmark \\

\bottomrule
\end{tabular}

\vspace{1mm}
\footnotesize{
From Scratch: generate maps without existing maps or sensor observations.;
Global Topology: generate globally consistent road networks;

Lane-level: support HD-map lane-level representations;
Large-scale: generate large-scale maps;

Controllable: support explicit user control over map generation;
Simulation-ready: directly deployable in driving simulators.

\cmark: fully supported;
\pmark: partially supported under restricted settings (e.g., predefined rules, local generation);
\xmark: not supported.
}
\end{table*}

Many driving simulators rely on maps that are manually designed and assembled.
CARLA and Highway-env, for example, provide handcrafted road environments for controlled and reproducible experiments~\cite{dosovitskiy2017carla,highway-env}.
Such maps are effective for benchmark development, but extending them to large road networks requires substantial engineering effort.
Their structural diversity is also limited by the available map assets and manual design process.

Another line of work constructs HD maps from real-world observations.
Traditional reconstruction pipelines recover lane geometry and semantic elements from vehicle-mounted or aerial sensor data~\cite{liao2025learning}.
Recent learning-based methods, including HDMapNet~\cite{li2022hdmapnet}, MapTR~\cite{liao2022maptr}, and VectorMapNet~\cite{liu2023vectormapnet}, predict vectorized map elements from multi-view images or LiDAR observations.
These approaches achieve high geometric fidelity but are restricted to existing road networks.
Their reliance on data collection and preprocessing also makes generating novel topologies and large-scale maps costly.

Rule-based generation methods combine road networks with predefined road elements under geometric or topological constraints. 
MetaDrive~\cite{li2022metadrive} assembles road primitives to create configurable driving environments, while RoadGen~\cite{yang2024roadgen} connects parameterized road components subject to geometric and connectivity rules.
These methods are efficient and offer direct control over map attributes.
Nevertheless, the generated structures remain constrained by manually specified primitives and composition rules.
Increasing map scale or topology diversity often requires more templates and rules, and the resulting lane geometry may lack the variation as in real maps.

Data-driven methods instead learn the distribution of road structures from existing maps.
HDMapGen~\cite{mi2021hdmapgen} models HD maps using a hierarchical graph generative framework, and ControlMap~\cite{farag2026controlmap} introduces spatial conditions to improve generation controllability. 
Compared with procedural approaches, learned models can capture richer structural patterns without explicitly encoding every road configuration.
However, existing methods are mainly designed for local map synthesis or bounded map patches.
They do not construct complete large-scale lane-level maps with consistent connectivity, nor do they ensure that the generated maps can be readily used in closed-loop driving simulation.

Overall, existing methods improve different aspects of map generation, but none simultaneously supports large-scale, simulation-ready HD map generation for comprehensive autonomous driving system evaluation.


\section{Method}\label{section:method}

\subsection{Overview}

\begin{figure*}
    \centering
    \includegraphics[width=\linewidth]{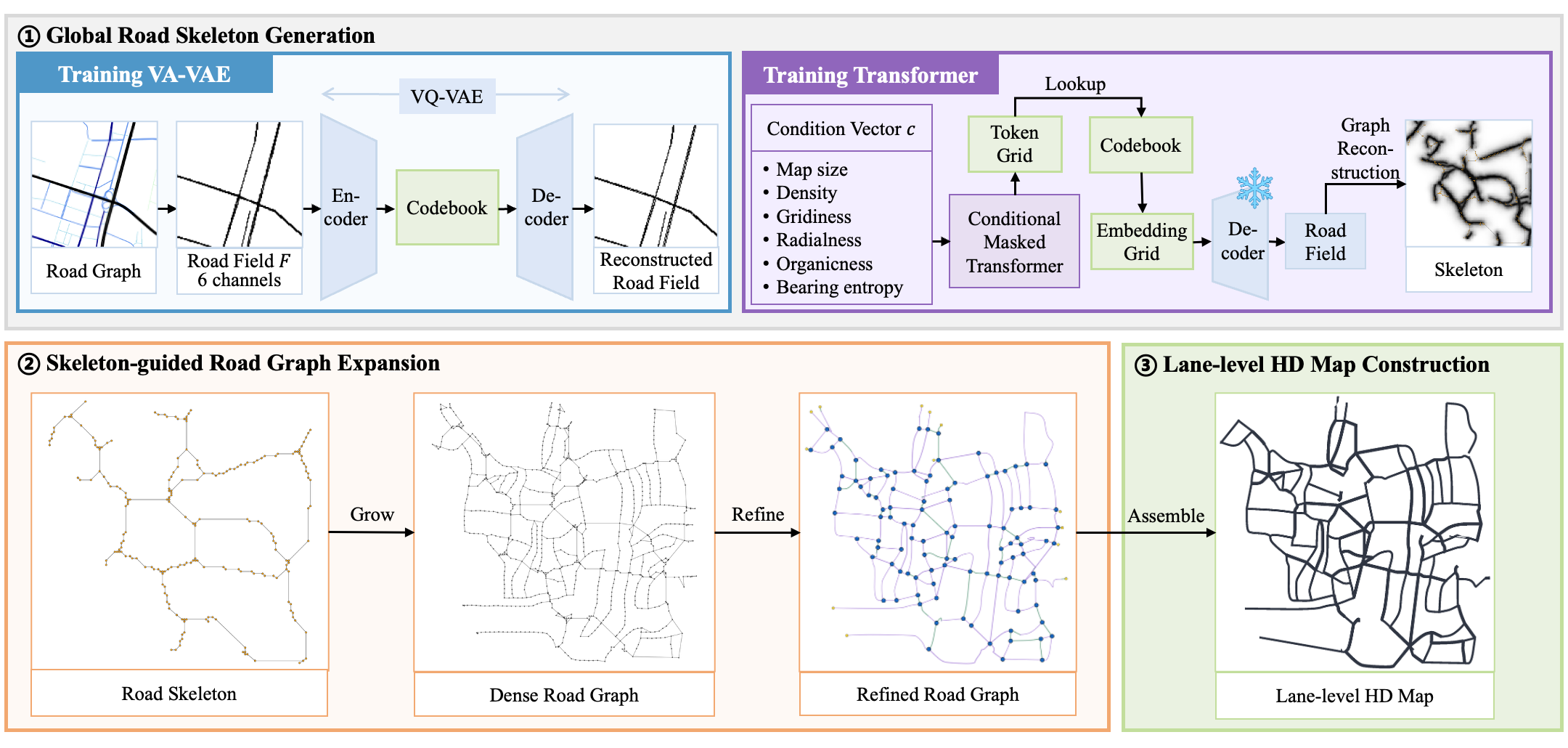}
    \caption{An overview of the map generation pipeline of RoadWeaver.}\label{fig:pipeline}
\end{figure*}

RoadWeaver adopts a coarse-to-fine generation strategy, in which a road network is constructed from global topology to lane-level geometry.
As illustrated in Fig.~\ref{fig:pipeline}, the proposed framework comprises three stages: global road skeleton generation, skeleton-guided road graph expansion, and topology-aware HD map construction.
The first stage generates a sparse road skeleton that captures the principal structure of the target road network.
Based on this skeleton, branch and local roads are subsequently introduced to improve geometric details.
The refined road graph is converted into a HD map while maintaining valid connectivity and topology.

RoadWeaver controls map scale by adjusting the spatial extent while maintaining a target road-network density.
By maintaining a target density while expanding the generation area, the number of generated road nodes can naturally increase without modifying the generation procedure.

\subsection{Global Road Skeleton Generation}\label{sec:skeleton}

Instead of generating all road segments simultaneously, we construct a sparse road skeleton that preserves the principal topology.
At this stage, geometric details are intentionally omitted, while long-range connectivity and the overall road layout are retained.
This decomposition reduces the complexity of the initial generation task and enables local roads to be added in subsequent stages without substantially changing the global network organization.

Since irregular road graphs cannot be directly represented on a regular spatial grid, we rasterize each graph into a road-field tensor
$\mathbf{F}\in\mathbb{R}^{H\times W\times 6}$, where $H$ and $W$ denote the spatial resolution.
Its six channels consist of one road probability channel, two orientation channels, one junction heatmap, one endpoint heatmap, and one distance field.
These channels preserve the geometric and topological cues for subsequent graph reconstruction.
During training, $\mathbf{F}$ is constructed from the ground-truth road graph extracted from OSM.

Because the objective of this stage is to model the global road topology, we represent the road field in a compact discrete latent space for simplicity and efficiency.
Specifically, we employ a VQ-VAE~\cite{van2017neural,esser2021taming} to encode $\mathbf{F}$ into a grid of discrete latent tokens and reconstruct the road field from the decoded tokens.
This discrete representation converts continuous field prediction into categorical token prediction, retaining the dominant topological structure while suppressing geometric details.
The VQ-VAE is optimized using reconstruction loss together with codebook and commitment losses.

To control the characteristics of the skeleton, we define a conditioning vector $\mathbf{c}\in\mathbb{R}^{11}$ consisting of a six-dimensional road-style code together with five structural priors: road density, gridness, radialness, organicness, and bearing entropy.
A conditional masked Transformer~\cite{chang2022maskgit} is trained to model the distribution of latent tokens conditioned on $\mathbf{c}$.
Road layouts exhibit strong long-range spatial dependencies, and the configuration of a local region depends on structures distributed across the entire map.
Masked token prediction can exploit bidirectional context and refine multiple regions in parallel, enabling more globally coherent layouts and more efficient sampling than autoregressive generation.
We use AdaLN conditioning~\cite{peebles2023scalable} to inject $\mathbf{c}$ into each Transformer block.

During inference, the sampled latent token grid is decoded by the VQ-VAE to recover the road field $\mathbf{F}$.
Road centerlines are then extracted from the decoded field and converted into a vectorized road skeleton through graph reconstruction.

\subsection{Skeleton-Guided Road Graph Expansion}

The generated skeleton specifies the main connectivity and large-scale organization of the road network, but does not provide sufficient local-road detail for lane-level map construction.
We expand the generated skeleton using a procedural road-growth strategy~\cite{parish2001procedural}.

Following the structure tensor formulation~\cite{bigun1987optimal}, we construct a continuous directional tensor field from tangent samples
collected along the generated skeleton.
For each spatial location $\mathbf{p}$, the tensor is defined as
\begin{equation}
\mathbf{T}(\mathbf{p})
=
\frac{
\sum_{i\in\mathcal{N}(\mathbf{p})}
g_i \mathbf{t}_i\mathbf{t}_i^{\top}
}{
\sum_{i\in\mathcal{N}(\mathbf{p})} g_i
}
\end{equation}
\begin{equation}
g_i
=
\exp\left(
-\frac{\|\mathbf{q}_i-\mathbf{p}\|_2^2}
{2\sigma^2}
\right).
\end{equation}
Here, $\mathbf{T}(\mathbf{p})\in\mathbb{R}^{2\times2}$ is a symmetric positive-semidefinite matrix,
$\mathbf{t}_i\in\mathbb{R}^{2}$ is the unit tangent direction associated with the $i$-th sampled skeleton point $\mathbf{q}_i$,
and $\mathcal{N}(\mathbf{p})$ denotes the local neighborhood of $\mathbf{p}$.
The coefficient $g_i$ represents the spatial contribution of each sample,
where $\sigma$ controls the decay bandwidth of the Gaussian kernel.
A separate neighborhood radius determines which samples are included in $\mathcal{N}(\mathbf{p})$.

Road growth is initialized from seed points sampled on both sides of the skeleton.
Let $t$ denote the current propagation step, where $\mathbf{p}_t\in\mathbb{R}^{2}$ and $\mathbf{d}_t\in\mathbb{R}^{2}$ denote the position and unit propagation direction of an active growth front, respectively.
We define
\begin{equation}
\mathbf{a}(\mathbf{p}_t,\mathbf{d}_t)
=
\underset{
\mathbf{v}\in
\left\{
\pm\mathbf{e}_{\mathrm{major}}(\mathbf{p}_t),
\pm\mathbf{e}_{\mathrm{minor}}(\mathbf{p}_t)
\right\}
}{\arg\max}
\;
\mathbf{v}^{\top}\mathbf{d}_t,
\label{eq:axis_selection}
\end{equation}
where
$\mathbf{e}_{\mathrm{major}}(\mathbf{p}_t)$
and
$\mathbf{e}_{\mathrm{minor}}(\mathbf{p}_t)$
are the major and minor eigenvectors of
$\mathbf{T}(\mathbf{p}_t)$,
respectively, and
$\mathbf{a}(\mathbf{p}_t,\mathbf{d}_t)$
denotes the eigen-direction of
$\mathbf{T}(\mathbf{p}_t)$
that is most closely aligned with the current propagation direction.

The deterministic candidate direction is computed as
\begin{equation}
\widetilde{\mathbf{d}}_{t+1}
=
\mathcal{C}_{\theta_{\max}}
\left(
\frac{
w_\mathrm{tensor} \mathbf{a}(\mathbf{p}_t,\mathbf{d}_t)
+
w_\mathrm{inertia} \mathbf{d}_t
}{
\|
w_\mathrm{tensor} \mathbf{a}(\mathbf{p}_t,\mathbf{d}_t)
+
w_\mathrm{inertia} \mathbf{d}_t
\|_2
}
\right),
\label{eq:direction_deterministic}
\end{equation}
where $w_\mathrm{tensor}$ and $w_\mathrm{inertia}$ are scalar weights for tensor guidance and directional inertia, respectively.
The operator $\mathcal{C}_{\theta_{\max}}$ limits the angular deviation from $\mathbf{d}_t$ to a prescribed maximum turning angle $\theta_{\max}$.

The final propagation direction $\mathbf d_{t+1}$ is obtained by applying a style-dependent angular perturbation $\Delta\phi$ to $\widetilde{\mathbf d}_{t+1}$.
The position of the growth front is then advanced using a first-order Euler step:
\begin{equation}
\mathbf{p}_{t+1} =
\mathbf{p}_t
+ \Delta s \,\mathbf{d}_{t+1},
\label{eq:euler}
\end{equation}
where $\Delta s$ denotes the propagation step length.

The propagation of an active front terminates when it leaves the map boundary, reaches the prescribed maximum length or iteration limit, or approaches an existing road within the snapping radius.
In the latter case, the newly generated road is attached to the existing graph.

After procedural growth, dangling endpoints are reconnected to nearby roads through A* search~\cite{hart1968formal} over a cost map derived from the decoded road field $\mathbf{F}$.
Regions with stronger predicted road evidence are assigned lower traversal costs, encouraging the reconnection paths to remain consistent with the spatial evidence predicted by the generative model.
Residual disconnected fragments are subsequently removed through largest-connected-component filtering.

Once the principal road graph has been formed, large unoccupied regions are detected and further populated with local roads.
These additional roads are generated using the same structure tensor field $\mathbf{T}$, which helps maintain directional consistency with the surrounding road network.

\subsection{Lane-Level HD Map Construction}

The final stage of RoadWeaver transforms the refined road graph into a lane-level HD map through a sequence of graph transformation and lane construction operations.
As summarized in Algorithm~\ref{alg:lane_hdmap}, the road graph is first refined through geometric regularization, topology simplification, and junction-aware processing.
Lane configurations $C$ are then assigned to individual road segments, from which lane boundaries $B$ and lane instances $L$ are generated.
Junction connector lanes are generated and lane-level topology is subsequently established to construct a directed lane graph.
Finally, geometric conflicts are repaired, the HD map is assembled, and a topology repair pass is applied to recover disconnected lane connections, producing the final lane-level HD map.

\begin{algorithm}[htb]
\caption{Lane-Level HD Map Construction Pipeline}\label{alg:lane_hdmap}
\begin{algorithmic}[1]

\Require Semantic road graph $G=(V,E)$ with geometric and road attributes
\Ensure Lane-level HD map $M$ with lane geometries and topology

\State $G \gets \Call{RefineRoadGraph}{G}$
\State $C=\{c_e\}_{e\in E}\gets\Call{AssignLaneConfig}{G}$

\For{each road segment $e\in E$}
    \State $B_e \gets \Call{GenerateLaneBoundaries}{e,c_e}$
    \State $L_e \gets \Call{ConstructDirectionalLanes}{B_e}$
\EndFor

\For{each junction $J$}
    \State $L_J^{con} \gets \Call{ConstructJunctionConnections}{J,L}$
\EndFor

\State $L \gets \Call{MergeLaneInstances}{\{L_e\},\{L_J^{con}\}}$

\State $L \gets \Call{AddSuccessorLinks}{L}$
\State $L \gets \Call{RepairLaneGeometry}{L}$
\State $M \gets \Call{AssembleHDMap}{L}$
\State $M \gets \Call{RepairTopology}{M}$
\State \Return $M$

\end{algorithmic}
\end{algorithm}


\section{Experiments}\label{section:experiments}

\subsection{Experiment Setup}

\subsubsection{Dataset}

We train RoadWeaver on real-world road networks collected from OSM, comprising approximately 58,000 samples from 144 cities.
Each urban road-network sample covers an area of approximately $2~\mathrm{km} \times 2~\mathrm{km}$.
We define road-node density as the number of road nodes per km$^2$ under a consistent graph extraction procedure and use it as a controllable measure of road density.
During inference, the generation area can be flexibly adjusted according to the desired simulation scale.
By maintaining the target road-node density, larger spatial regions naturally result in larger road networks with more nodes while preserving consistent structural characteristics.

\subsubsection{Baselines}

We compare RoadWeaver with representative rule-based and data-driven map generation methods.
MetaDrive~\cite{li2022metadrive} and RoadGen~\cite{yang2024roadgen} are adopted as rule-based baselines, while HDMapGen~\cite{mi2021hdmapgen} is included as a data-driven baseline for local lane-level map generation.

\subsubsection{Hardware}

All experiments are conducted on a workstation equipped with an NVIDIA GeForce RTX 5090 GPU, an AMD Ryzen 7 9800X3D CPU, and 64~GB of system RAM. 
The reported generation time covers the complete map-generation pipeline and is averaged over multiple independent runs, excluding model-loading and warm-up overhead.

\subsection{Generation Quality, Diversity, and Controllability}

We first compare the generation quality of RoadWeaver with representative existing methods under a comparable map scale.
All methods are configured to generate road graphs containing approximately 35--40 nodes.

\begin{figure*}[htb]
    \centering
    \subfloat[HDMapGen (38 graph nodes)]{\includegraphics[height=4.7cm]{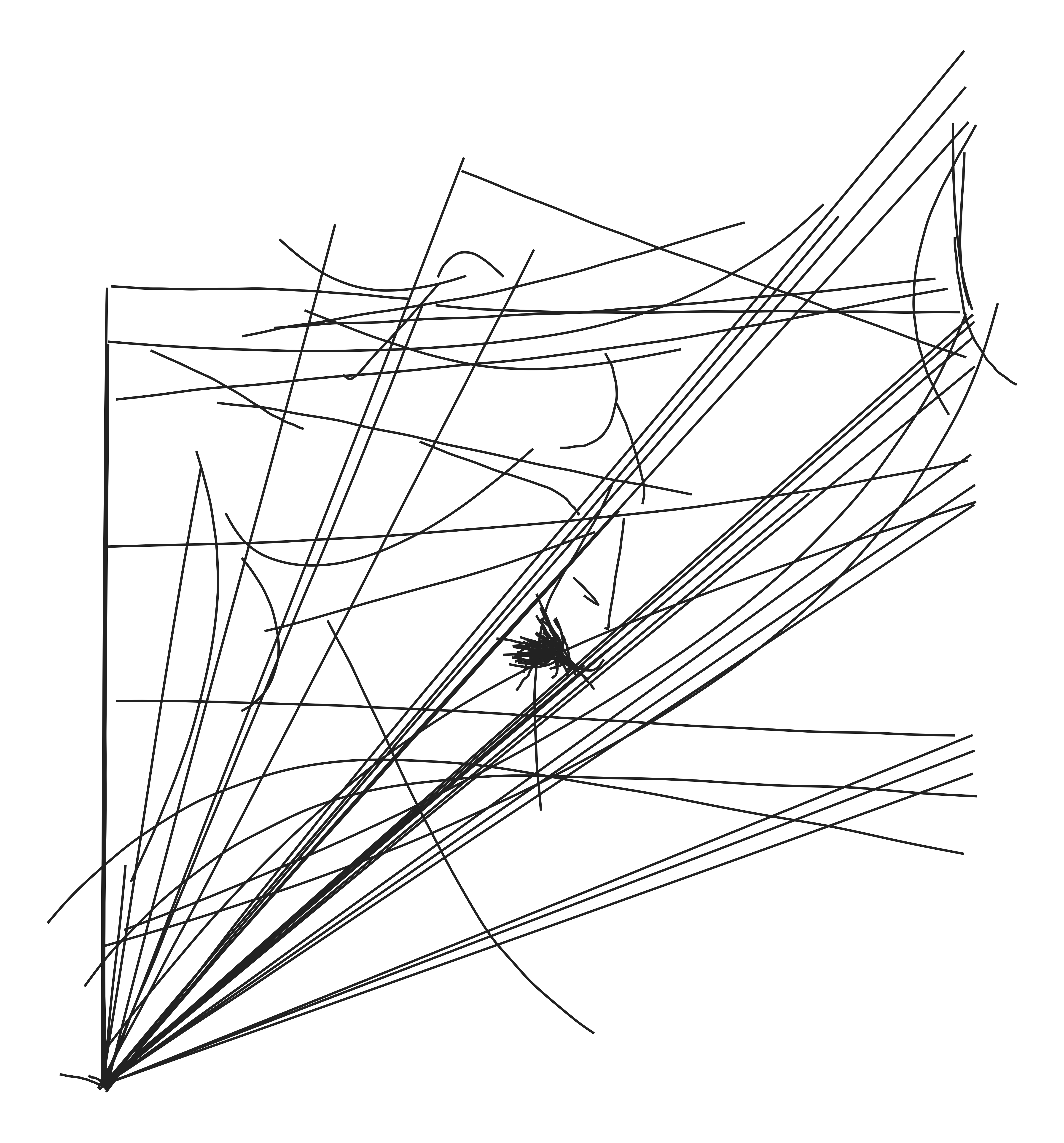}}
    \hfill
    \subfloat[MetaDrive (40 graph nodes)]{\includegraphics[height=4.7cm]{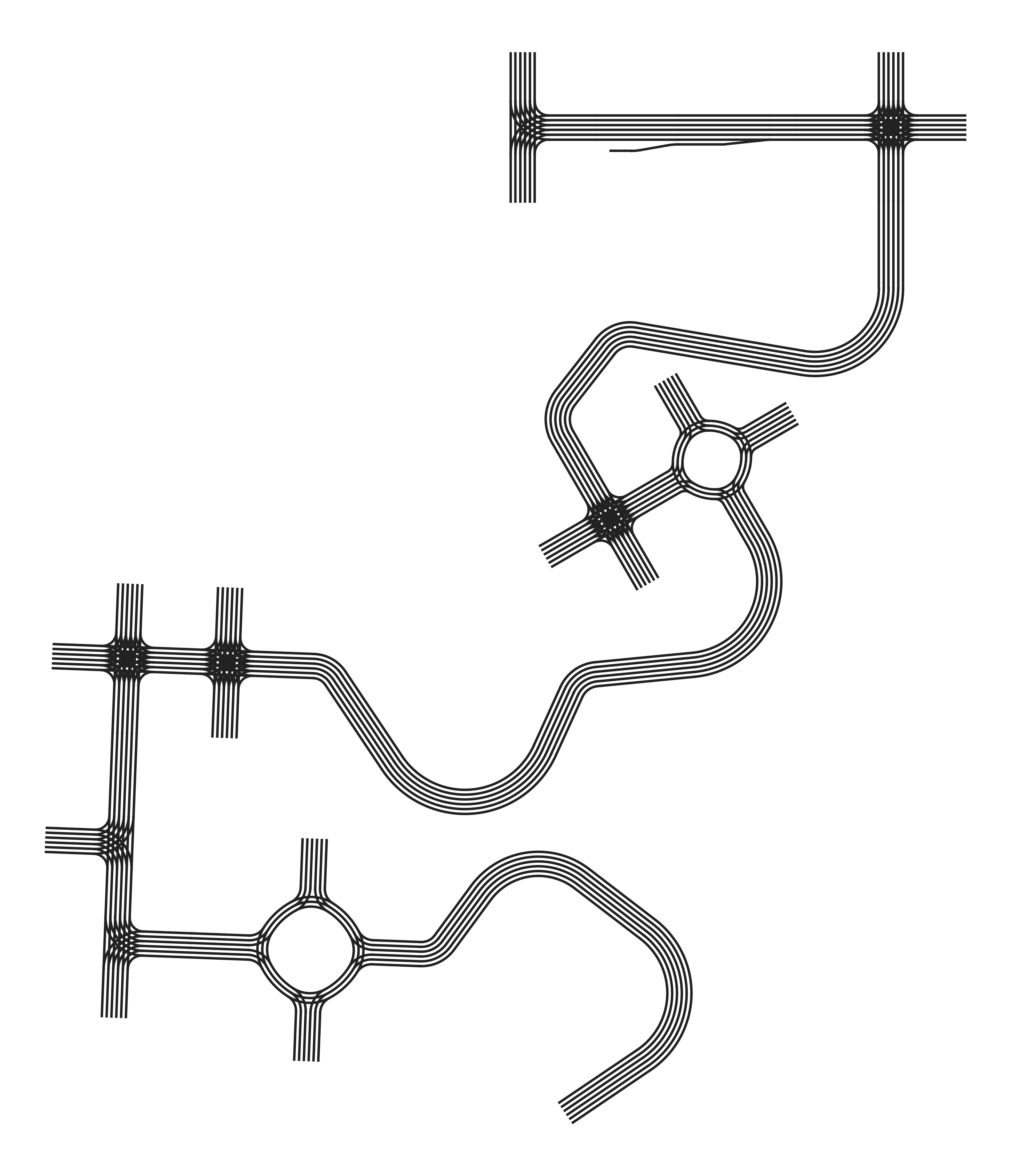}}
    \hfill
    \subfloat[RoadGen (37 graph nodes)]{\includegraphics[height=4.7cm]{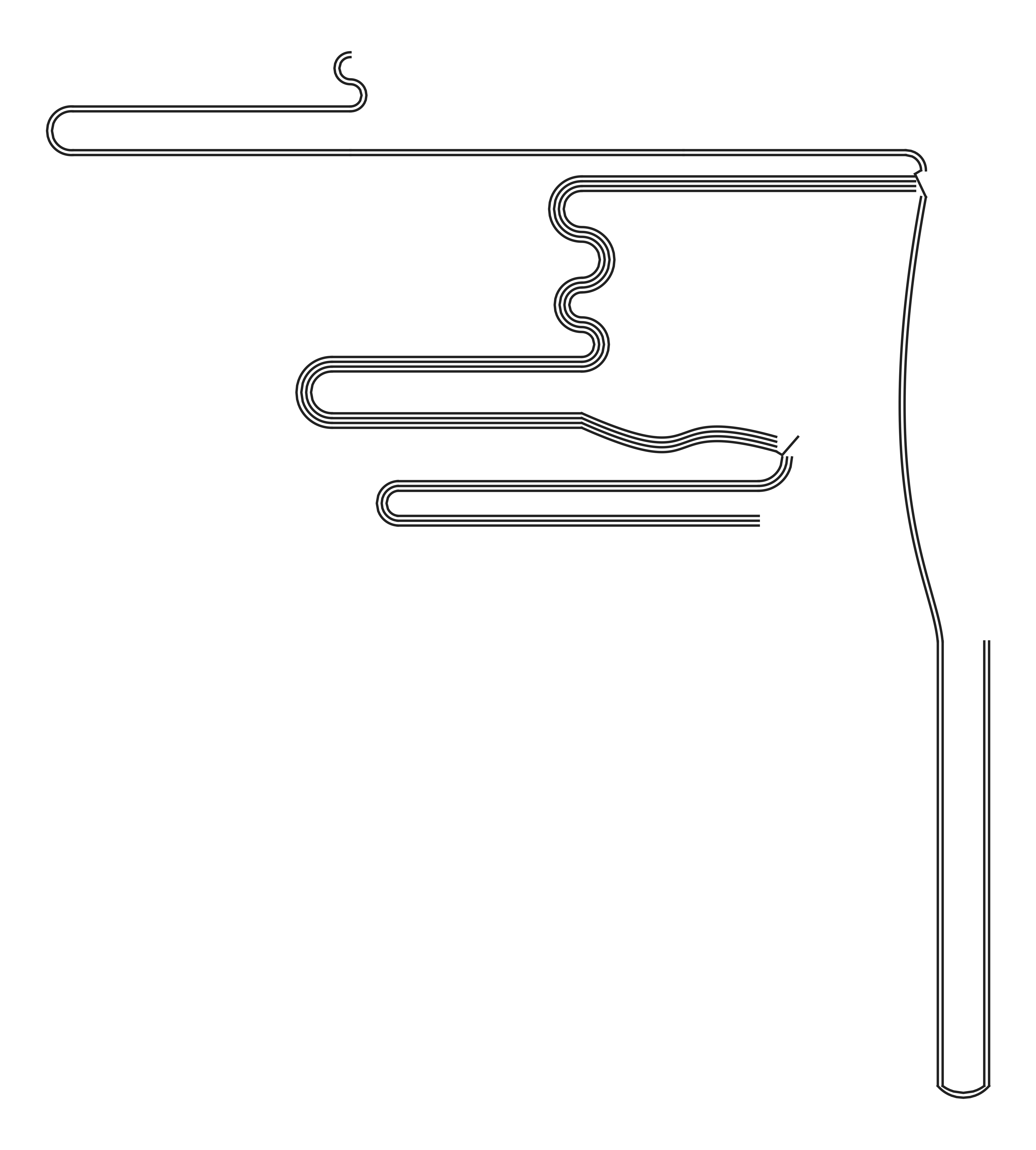}}
    \hfill
    \subfloat[RoadWeaver (40 graph nodes)]{\includegraphics[height=4.7cm]{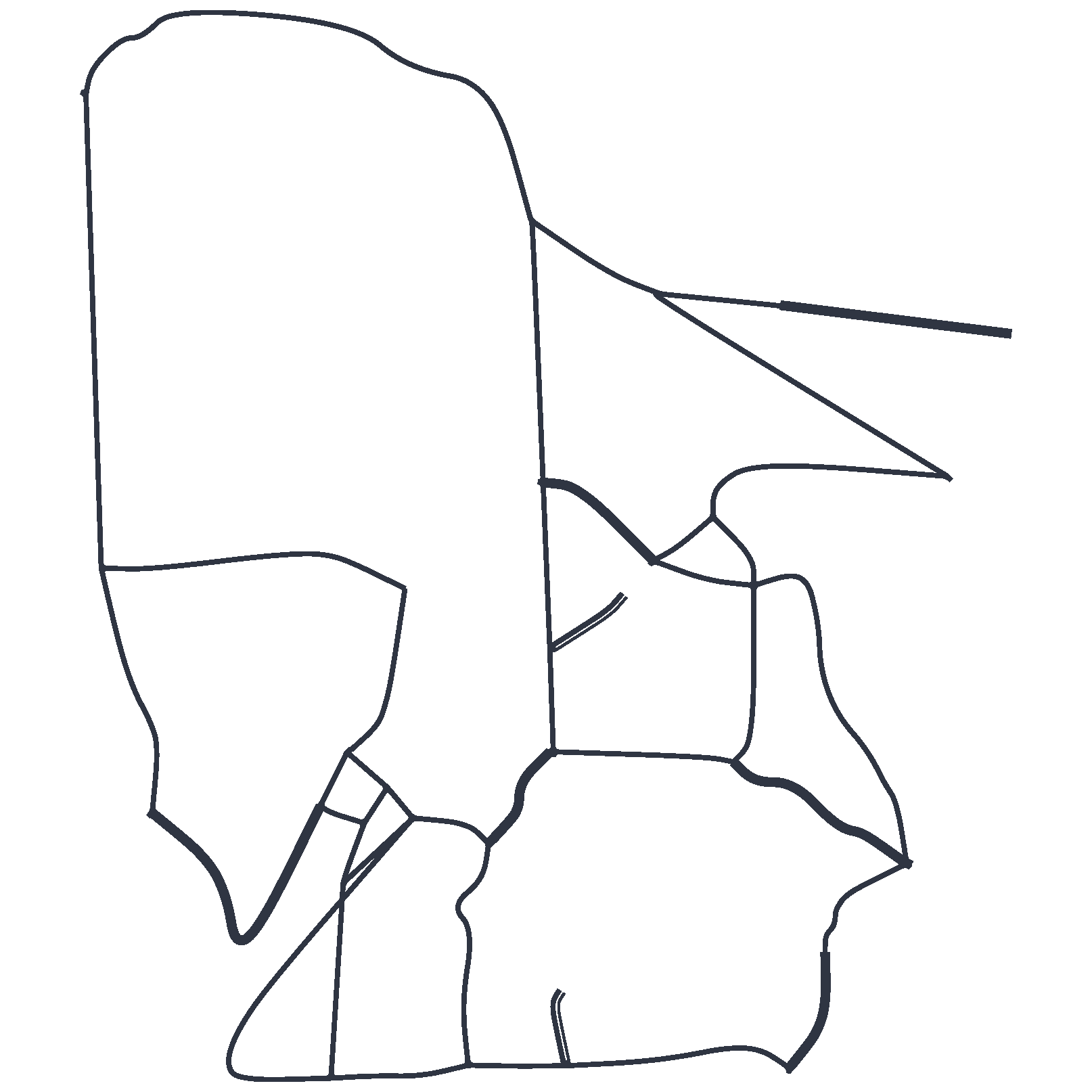}}
    \caption{Representative maps generated by different methods at a comparable map scale (35--40 graph nodes). Intersections, road endpoints, and turning transition points are counted as graph nodes.}\label{fig:quality}
\end{figure*}

As shown in Fig.~\ref{fig:quality}, RoadWeaver generates road networks with coherent topology and geometry across different regions of the map.
Specifically, RoadWeaver maintains connectivity among road segments while avoiding disconnected components and abrupt transitions observed in baseline methods.
The generated networks also preserve lane-level structures and intersection configurations, demonstrating the ability to jointly model road topology and geometric layouts.

\begin{figure*}[htb]
    \centering
    \subfloat[Target: 10; \\Actual: 8.9]{\includegraphics[width=0.12\linewidth]{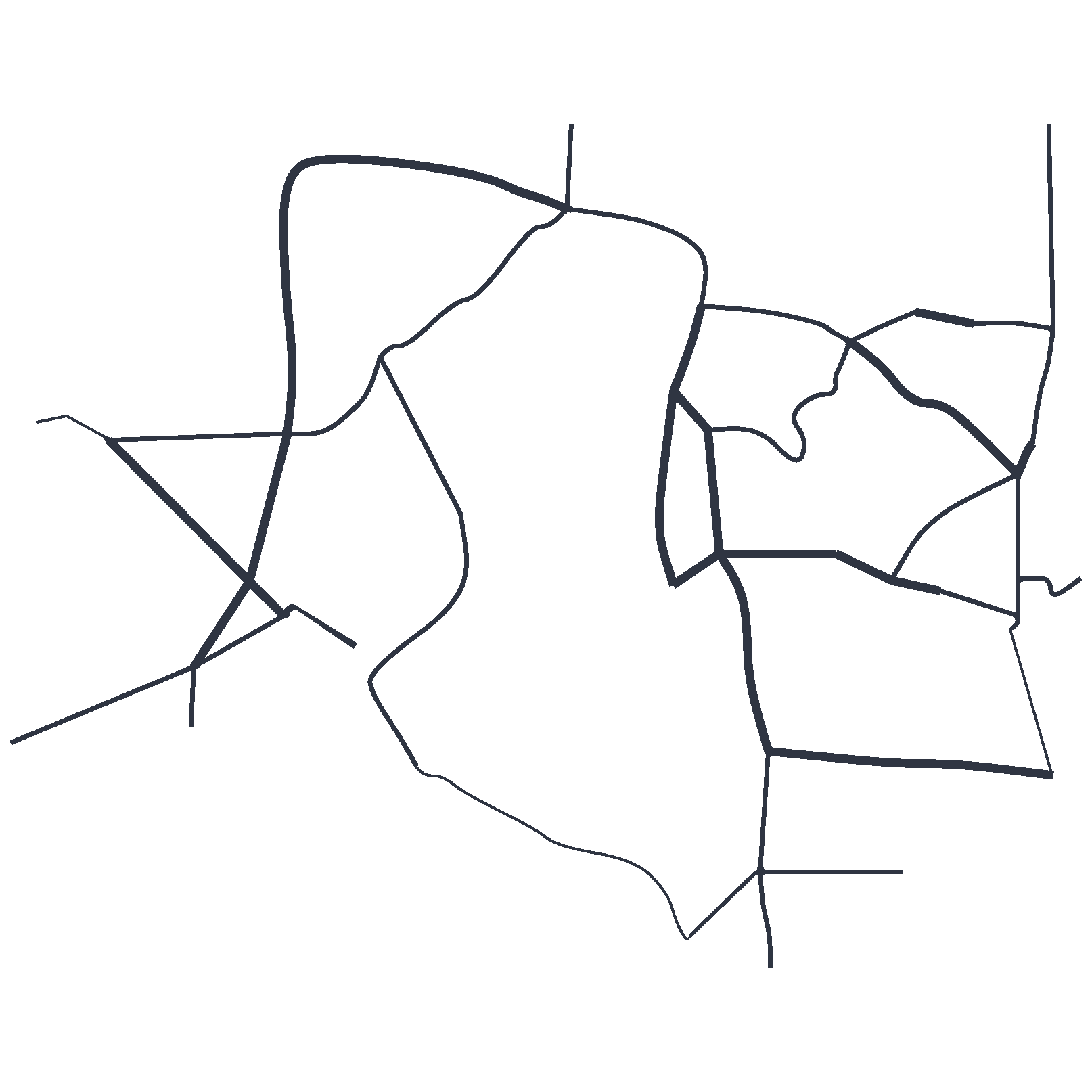}}
    \hfill
    \subfloat[Target: 15; \\Actual: 15.0]{\includegraphics[width=0.12\linewidth]{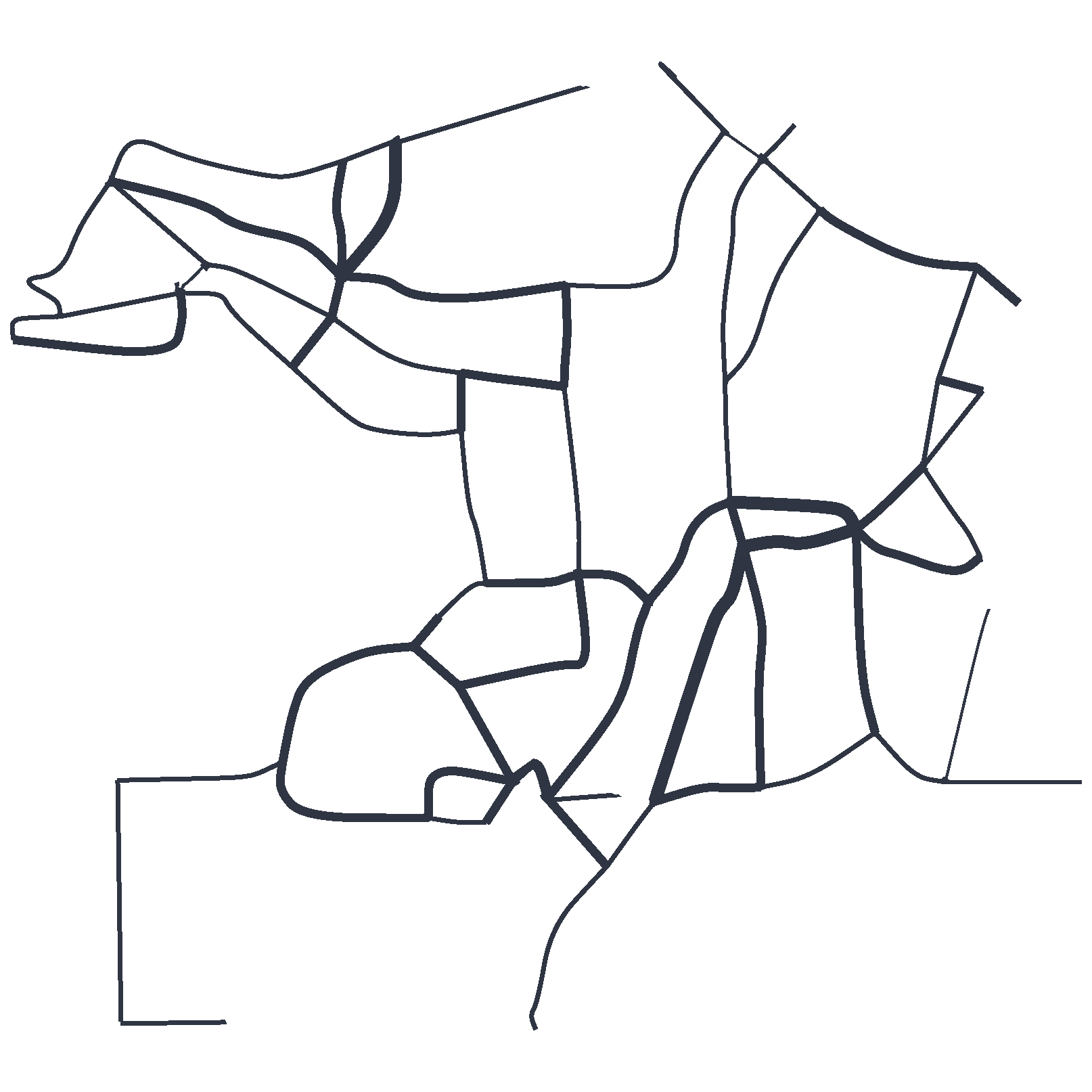}}
    \hfill
    \subfloat[Target: 20; \\Actual: 19.6]{\includegraphics[width=0.12\linewidth]{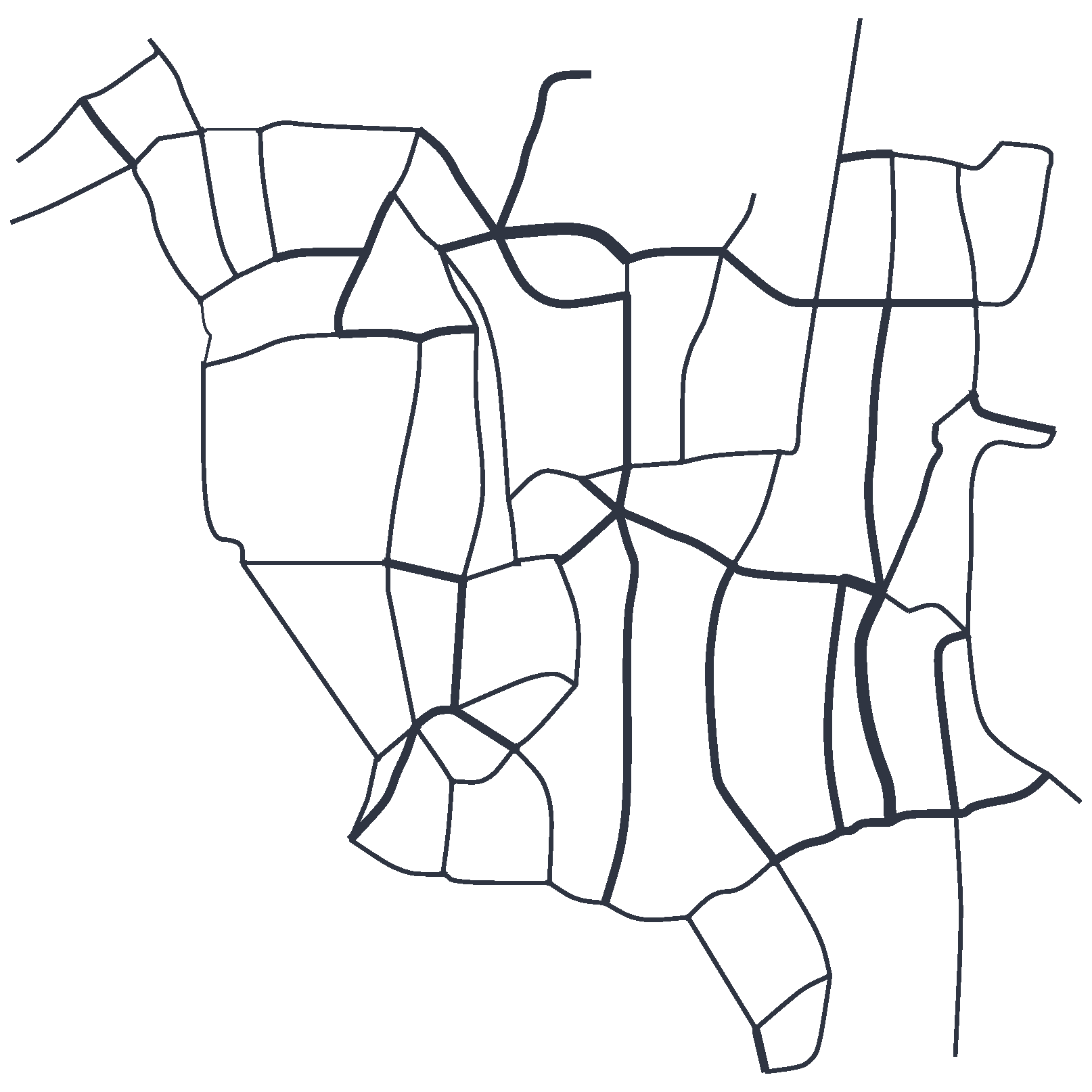}}
    \hfill
    \subfloat[Target: 25; \\Actual: 25.1]{\includegraphics[width=0.12\linewidth]{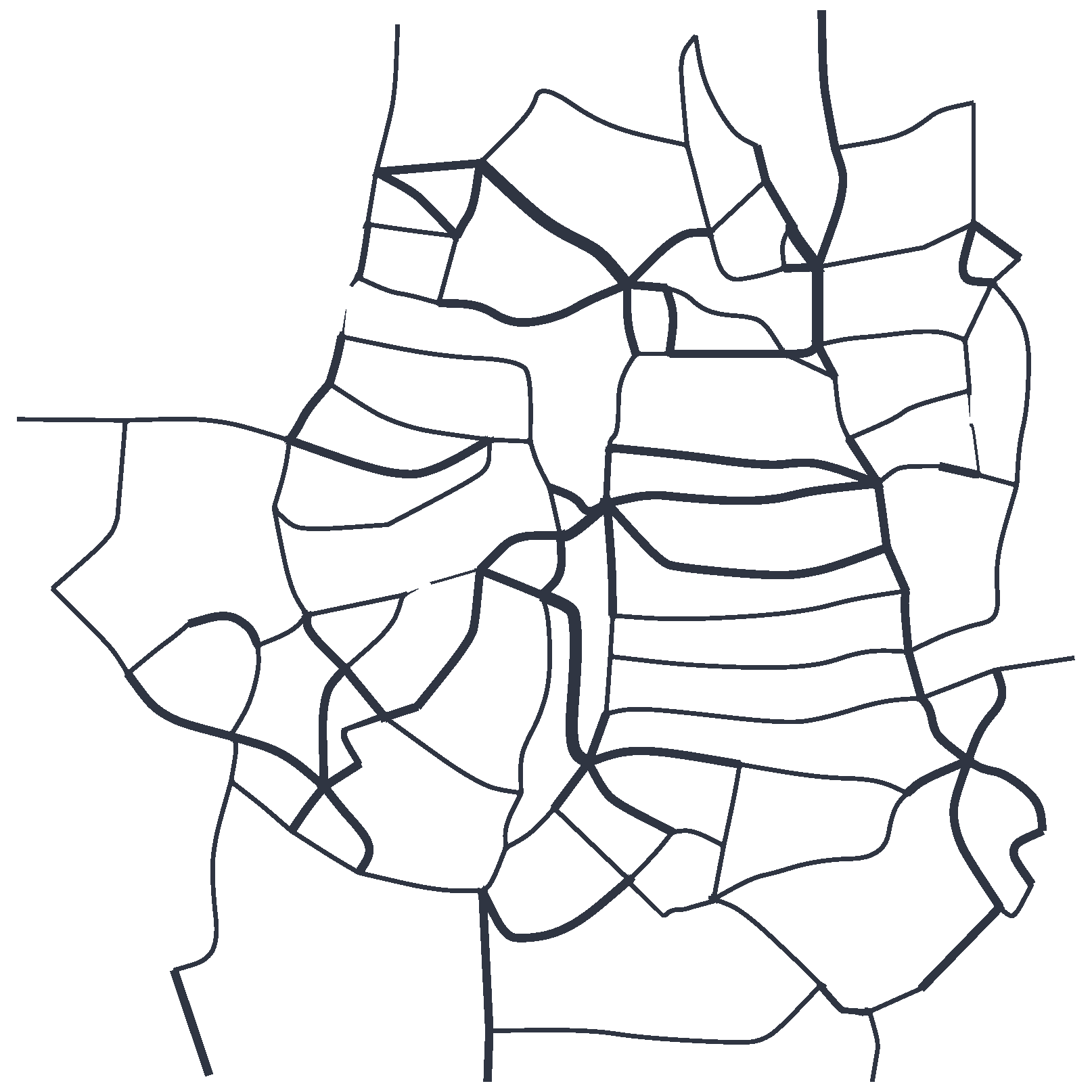}}
    \hfill
    \subfloat[Target: 30; \\Actual: 30.5]{\includegraphics[width=0.12\linewidth]{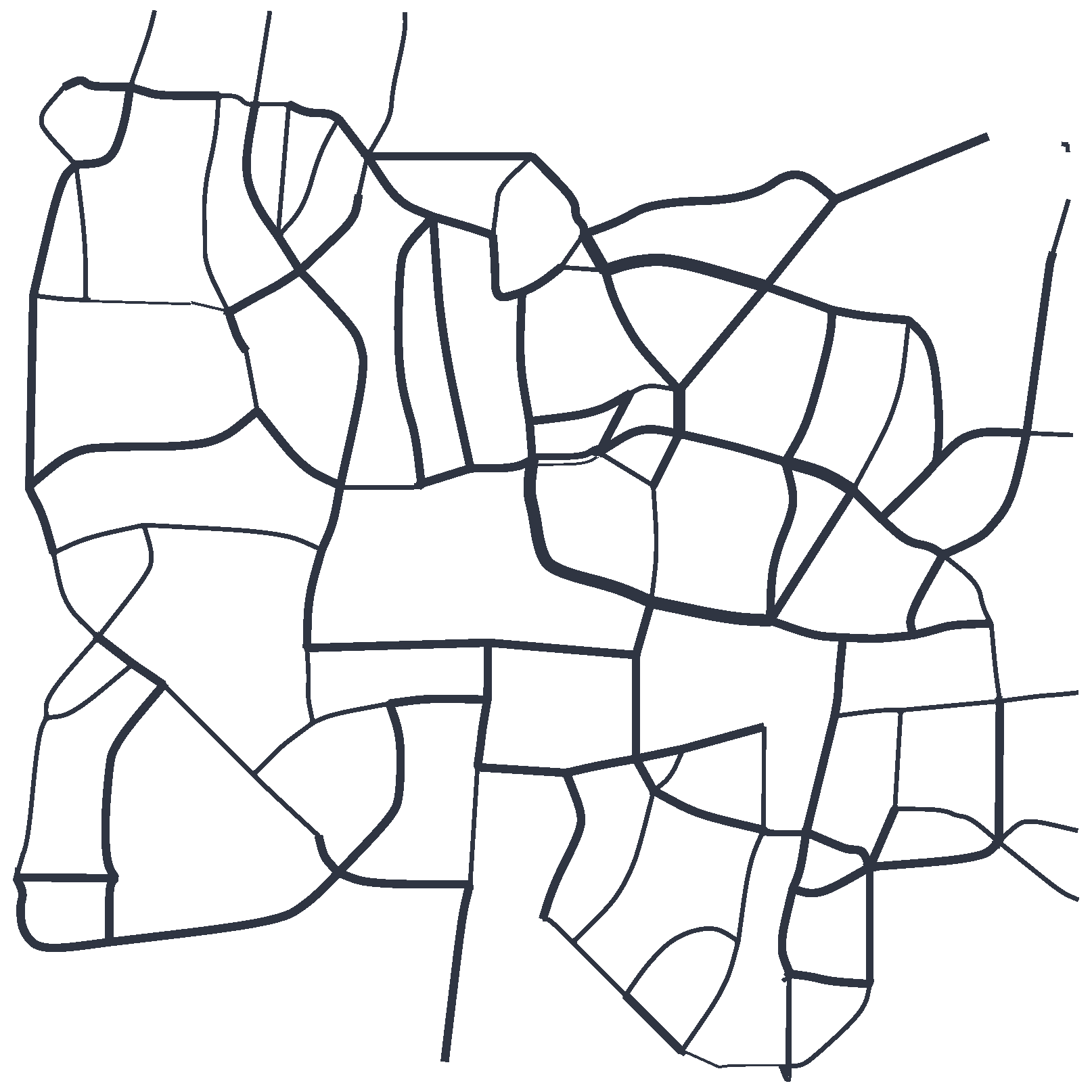}}
    \hfill
    \subfloat[Target: 35; \\Actual: 34.5]{\includegraphics[width=0.12\linewidth]{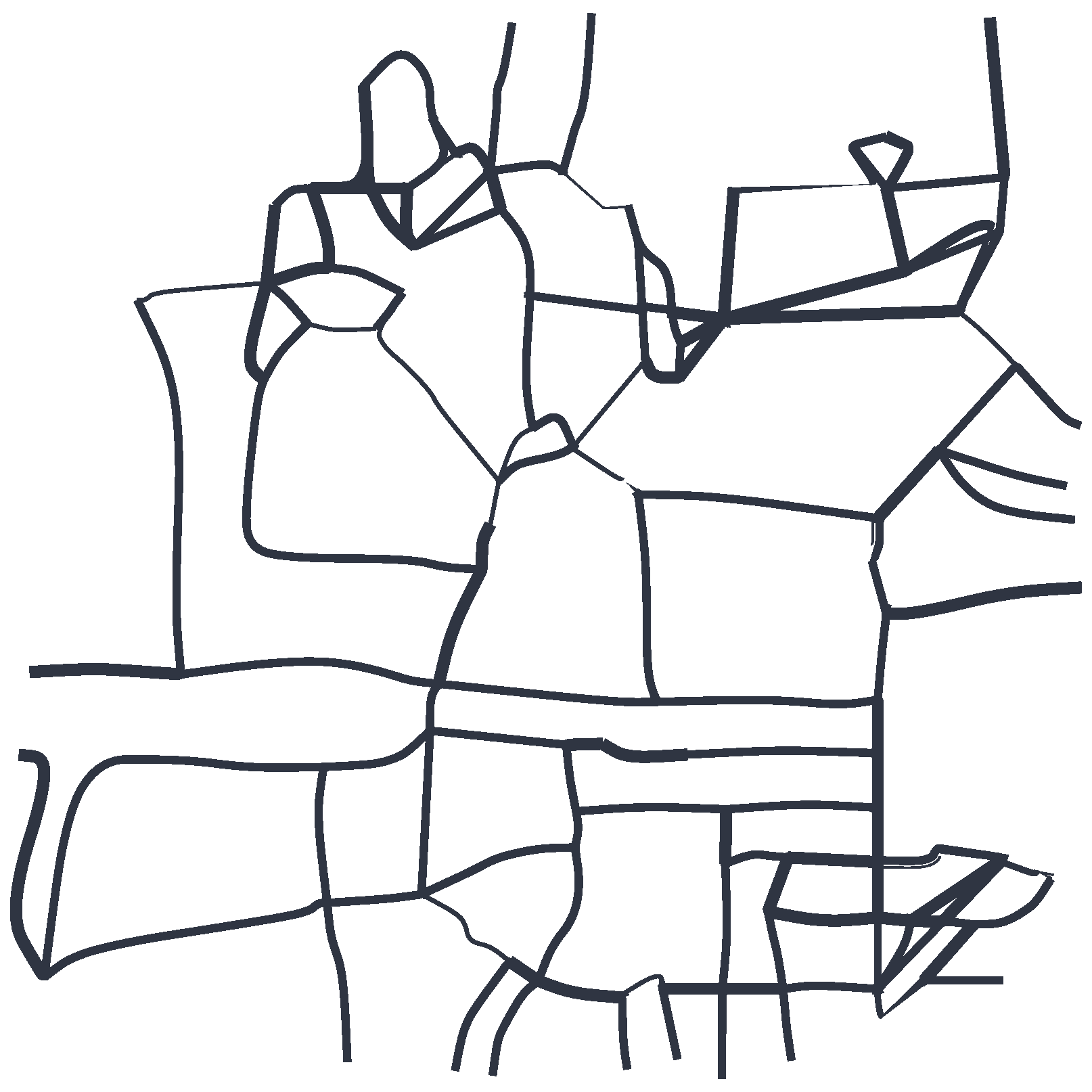}}
    \hfill
    \subfloat[Target: 40; \\Actual: 39.0]{\includegraphics[width=0.12\linewidth]{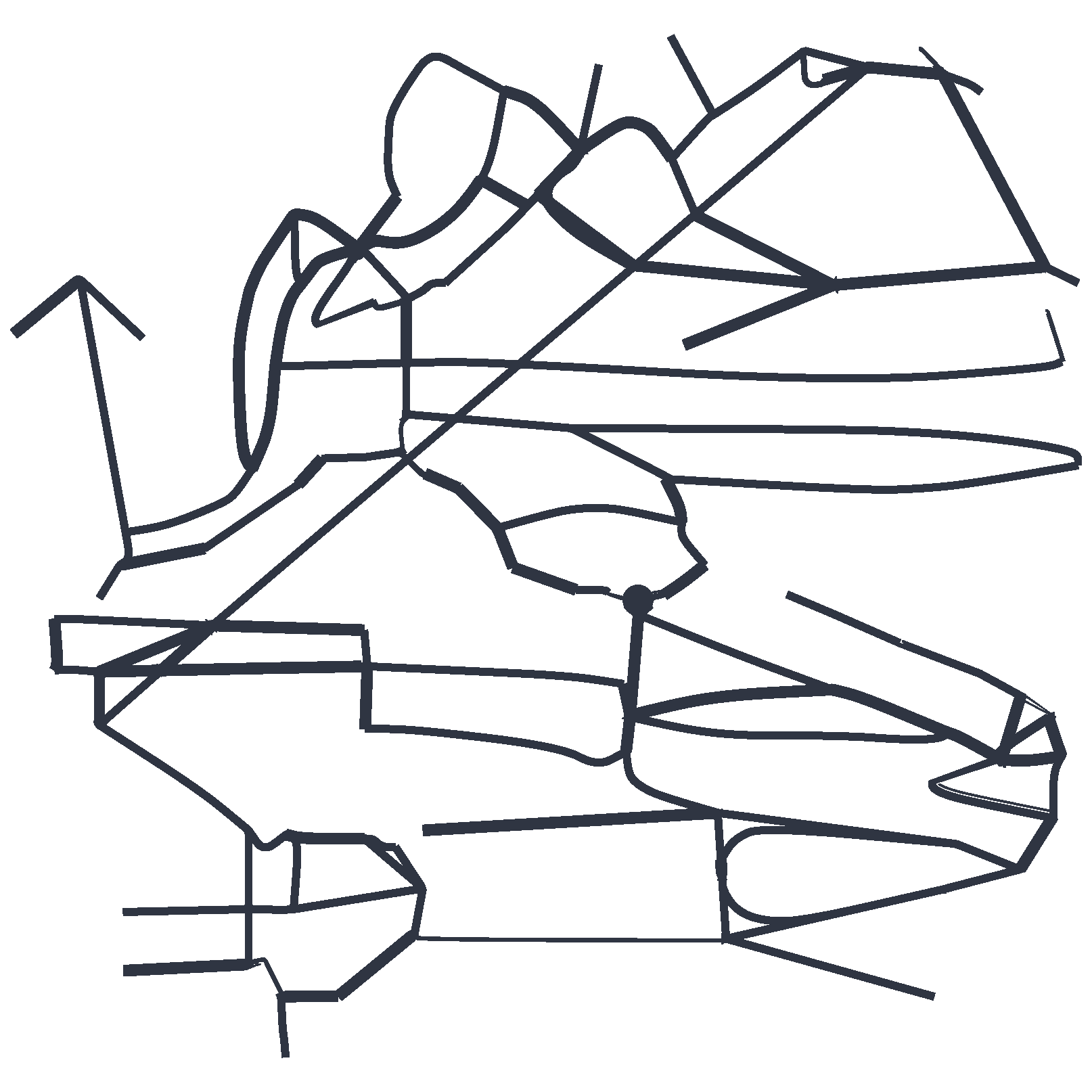}} \\
    \subfloat[Target: 10; \\Actual: 12.5]{\includegraphics[width=0.12\linewidth]{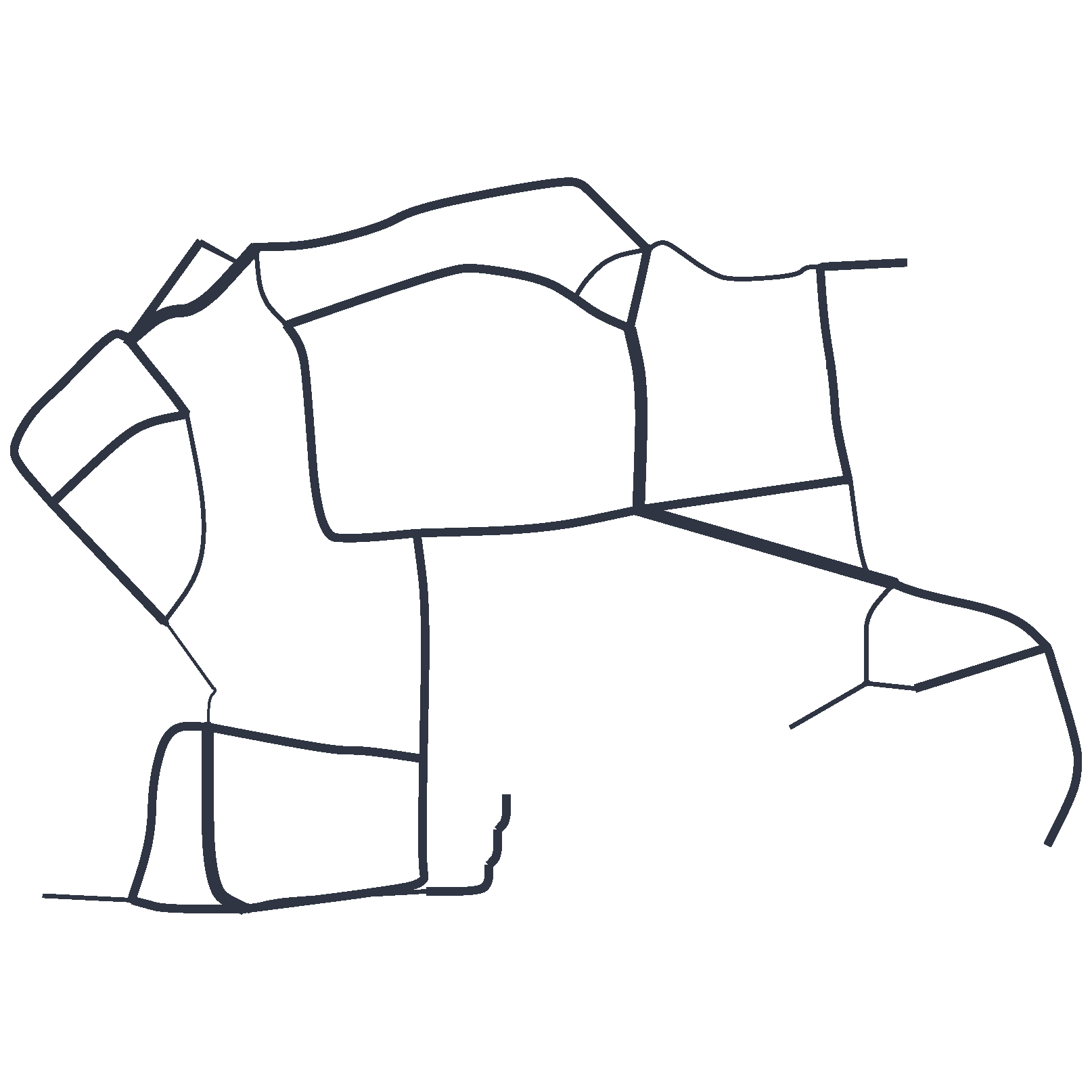}}
    \hfill
    \subfloat[Target: 15; \\Actual: 17.6]{\includegraphics[width=0.12\linewidth]{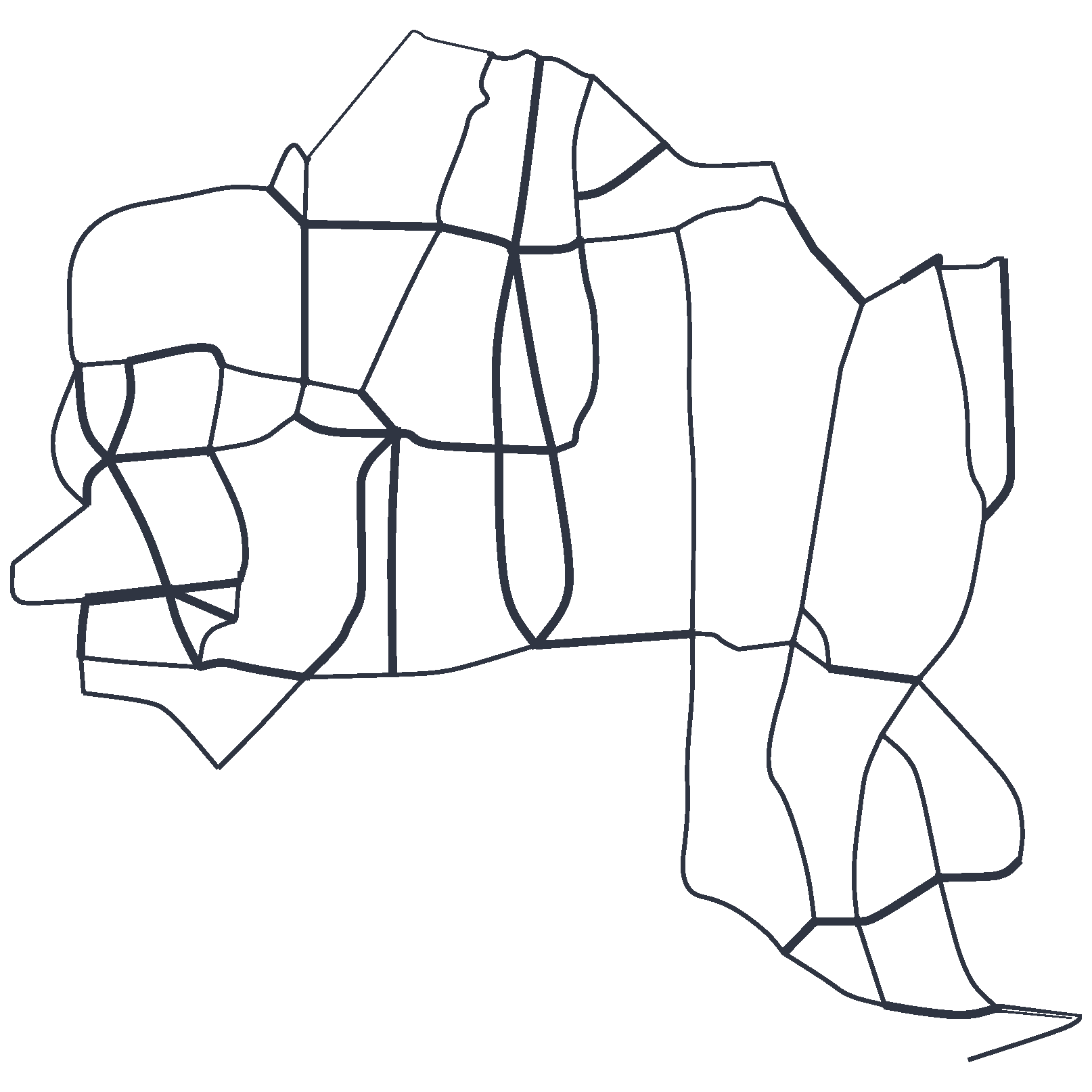}}
    \hfill
    \subfloat[Target: 20; \\Actual: 22.4]{\includegraphics[width=0.12\linewidth]{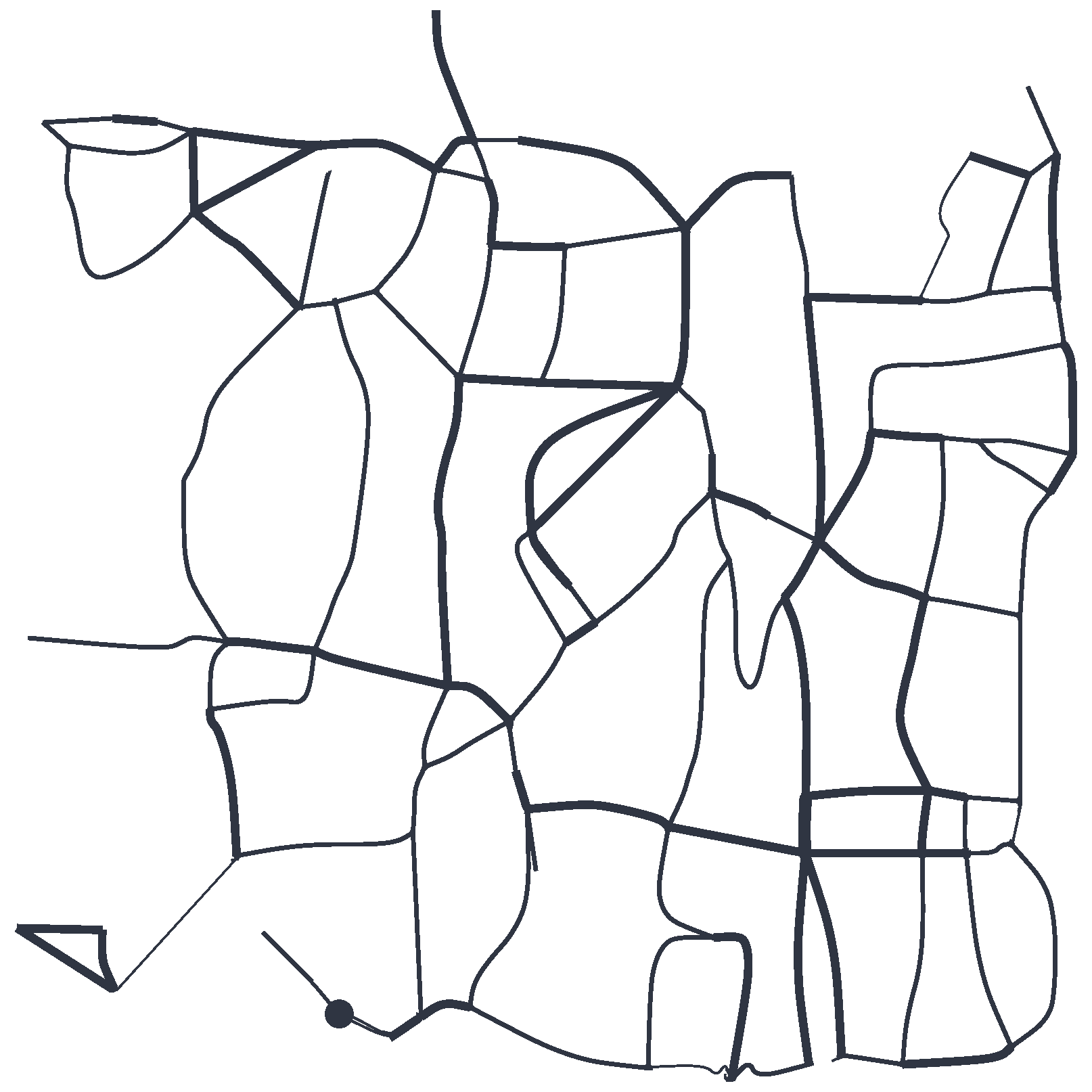}}
    \hfill
    \subfloat[Target: 25; \\Actual: 25.2]{\includegraphics[width=0.12\linewidth]{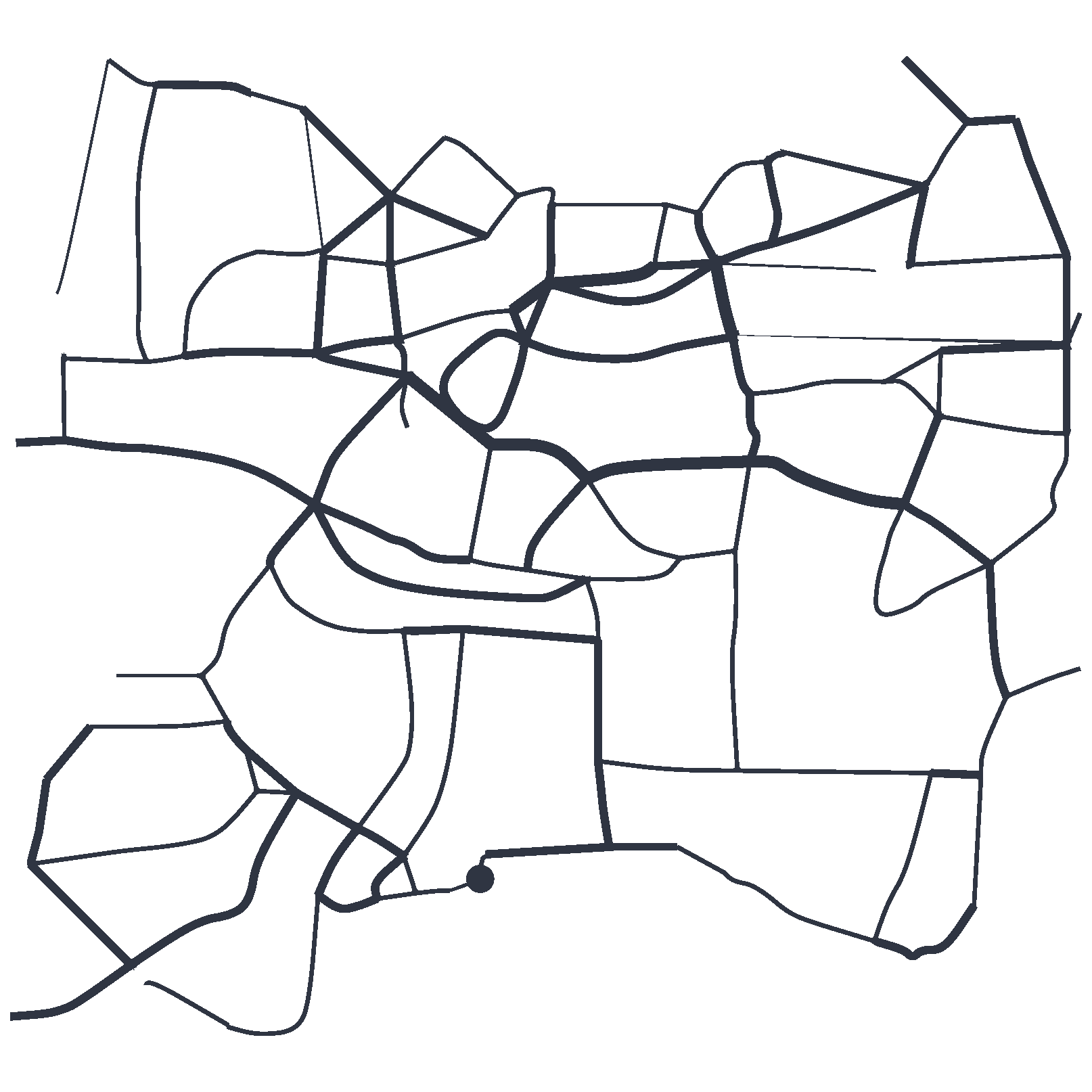}}
    \hfill
    \subfloat[Target: 30; \\Actual: 31.7]{\includegraphics[width=0.12\linewidth]{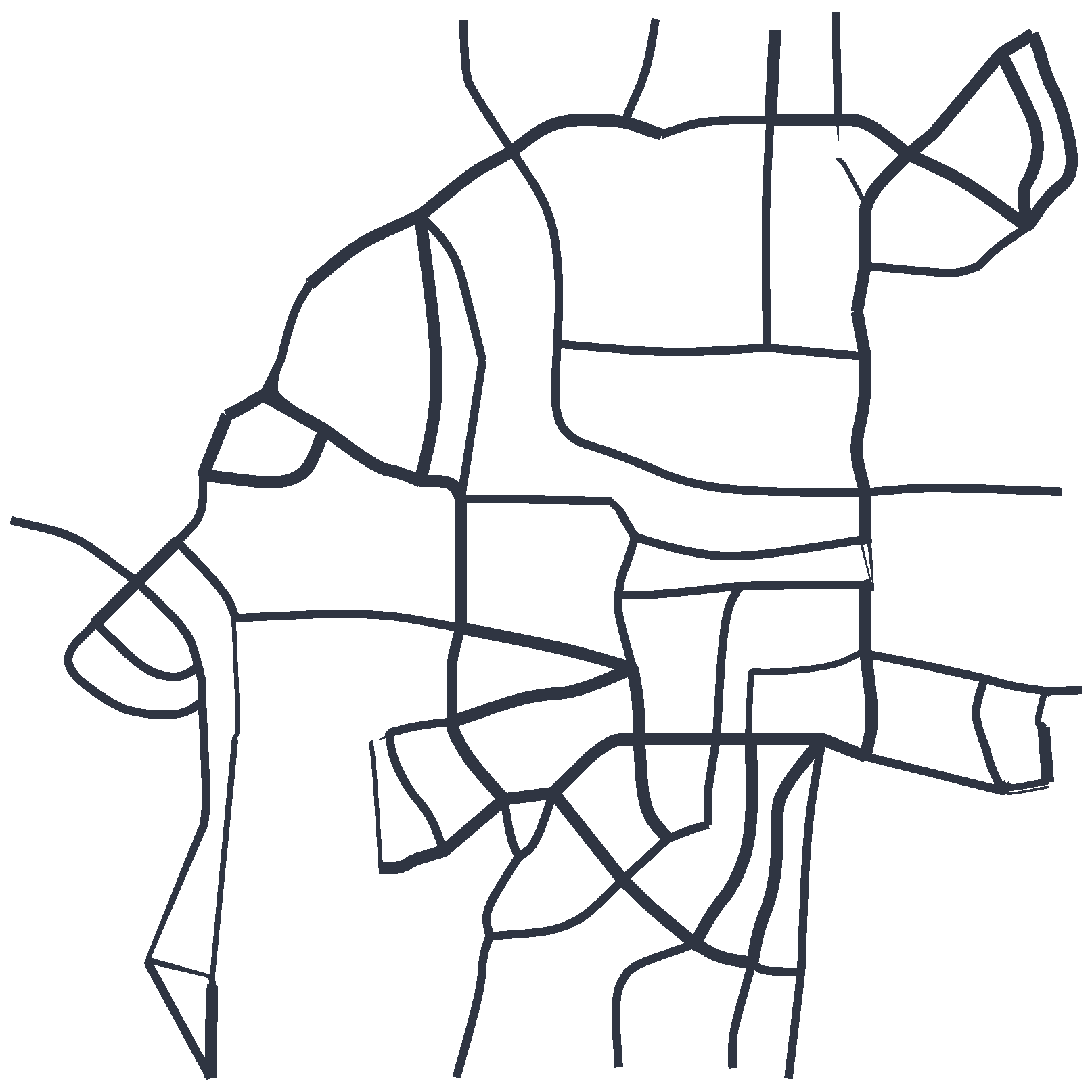}}
    \hfill
    \subfloat[Target: 35; \\Actual: 37.1]{\includegraphics[width=0.12\linewidth]{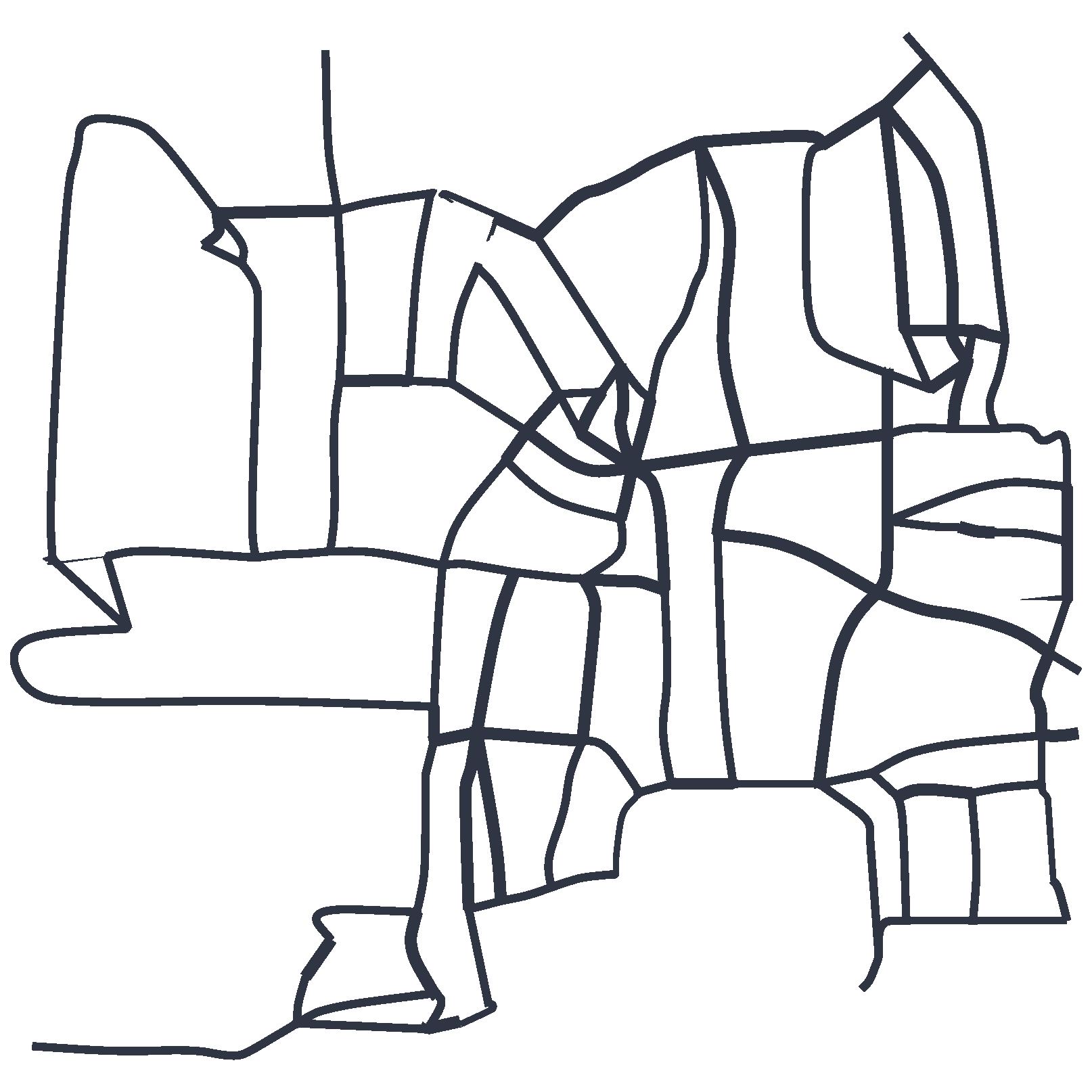}}
    \hfill
    \subfloat[Target: 40; \\Actual: 43.3]{\includegraphics[width=0.12\linewidth]{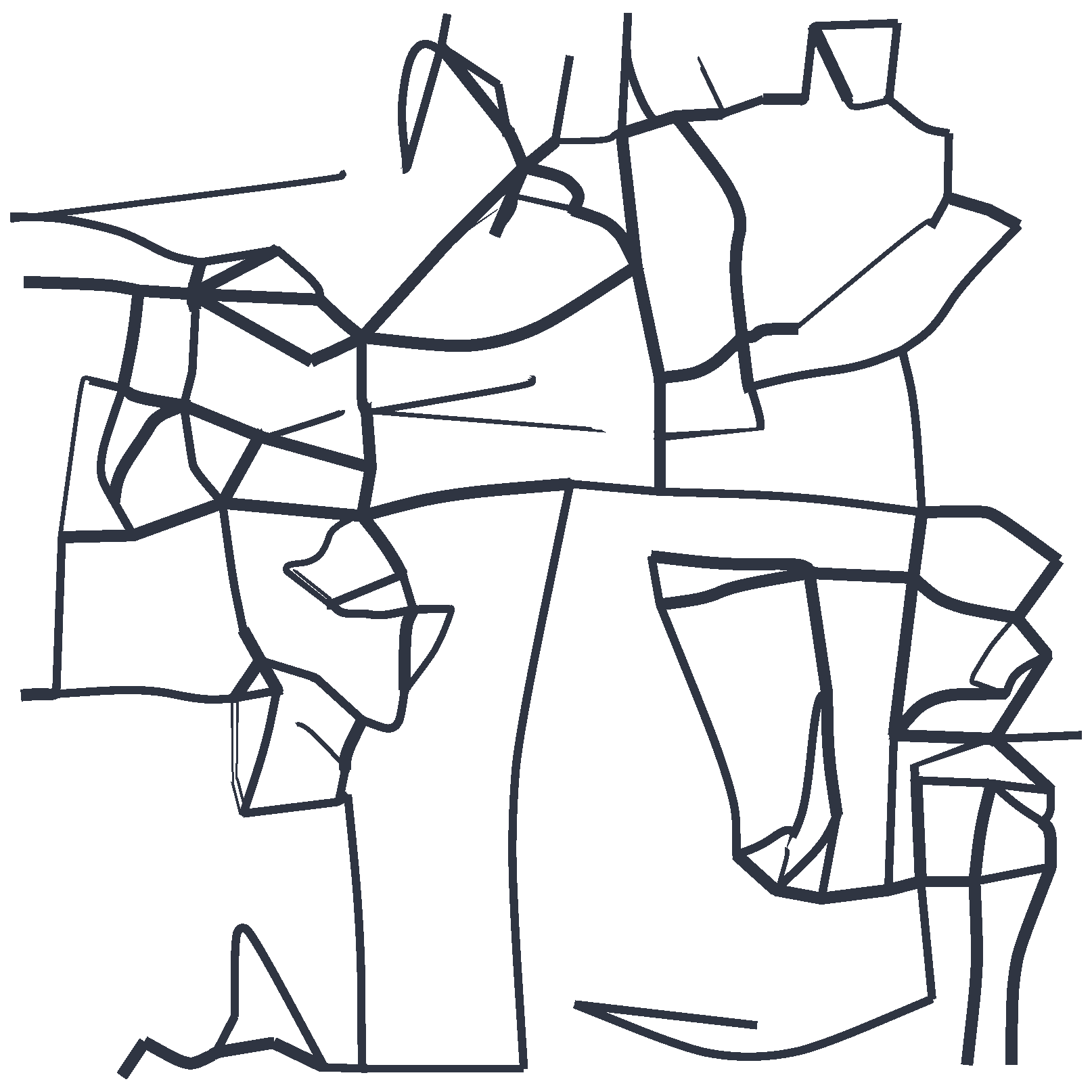}}
    \caption{Demonstration of RoadWeaver's controllability and diversity under different road-node density conditions. The road-node density is defined as the number of road nodes per km$^2$.}\label{fig:diversity-density}
\end{figure*}

Beyond generation quality, Fig.~\ref{fig:diversity-density} illustrates maps generated with different random seeds under the same target density.
The paired examples show $2~\mathrm{km}\times2~\mathrm{km}$ maps generated with target densities of 10, 15, 20, 25, 30, 35, and 40 road nodes/km$^2$.
Despite having similar measured densities, these maps exhibit distinct topological organizations, intersection configurations, and local geometric patterns.
These results demonstrate that RoadWeaver can generate varied road structures under the same density constraint while preserving consistent network-level statistics.

The same figure also demonstrates the controllability of RoadWeaver over road-network density.
By increasing the target density, the generated maps become progressively denser while maintaining connectivity and lane-level consistency.
The measured road-node densities remain close to the specified values, confirming that the density constraint can be effectively reflected in the generated networks.
When the target density exceeds approximately 40 road nodes/km$^2$, the density increase becomes less significant due to the increasing topological constraints of dense road networks.

\subsection{Topological and Geometric Validity}

Following previous road-network generation and HD map modeling studies~\cite{mi2021hdmapgen,yang2024roadgen,liu2023vectormapnet,liao2022maptr}, we evaluate the generated maps from topological and geometric perspectives using five metrics.
Specifically, the LCC ratio and reachability assess graph connectivity, the dead-end ratio and cycle ratio characterize network structure, and the endpoint alignment error evaluates lane-level geometric consistency.
The LCC ratio denotes the proportion of nodes in the largest connected component~\cite{boeing2017osmnx}.
Reachability is the fraction of node pairs connected by valid paths~\cite{boeing2017osmnx,mi2021hdmapgen,yang2024roadgen}.
The dead-end ratio quantifies the proportion of terminal nodes~\cite{boeing2017osmnx}.
The cycle ratio measures the proportion of nodes belonging to at least one cycle~\cite{boeing2017osmnx}.
The endpoint alignment error is defined as the average Euclidean distance between lane endpoints that are expected to be connected~\cite{liu2023vectormapnet,liao2022maptr}.

\begin{table*}[htb]
\centering
\caption{Topological and geometric validity of generated road networks.}
\label{table:validity}

\begin{tabular}{l|ccccc}
\toprule
\multirow{2}{*}{\textbf{Method}}
& \textbf{LCC} 
& \textbf{Reachability}
& \textbf{Dead-end Ratio}
& \textbf{Cycle Ratio}
& \textbf{Endpoint Alignment Error} \\
& (\%) $\uparrow$
& (\%) $\uparrow$
& (\%)  $\downarrow$
& (\%) $\uparrow$
& (m)  $\downarrow$ \\
\midrule
MetaDrive 
& $99.5 \pm 2.3$ 
& $99.3 \pm 3.3$
& $43.0 \pm 10.1$
& $50.5 \pm 13.8$
& $8.17 \pm 2.43$ \\
RoadGen 
& $\textbf{100.0} \pm 0.0$ 
& $\textbf{100.0} \pm 0.0$ 
& $12.8 \pm 6.2$
& $0.0 \pm 0.0$
& $4.80 \pm 1.29$ \\
HDMapGen 
& $64.3 \pm 24.7$ 
& $56.2 \pm 28.2$
& $42.9 \pm 29.4$
& $46.5 \pm 37.2$
& $4.32 \pm 0.41$ \\
\midrule
\textbf{RoadWeaver} 
& $99.9 \pm 0.3$ 
& $99.8 \pm 0.7$ 
& $\textbf{10.7} \pm 3.6$
& $\textbf{85.2} \pm 6.0$
& $\textbf{0.24} \pm 0.22$ \\
\bottomrule
\end{tabular}
\end{table*}

Table~\ref{table:validity} summarizes the results.
RoadWeaver achieves an LCC of $99.9\%$ and a reachability of $99.8\%$, showing that most generated road segments belong to a connected road network.
It also obtains the lowest dead-end ratio among all compared methods, indicating fewer disconnected road extensions.

The cycle ratio shows a clear difference among the generated graph structures.
RoadWeaver reaches $85.2\%$, compared with $50.5\%$ for MetaDrive, $46.5\%$ for HDMapGen, and $0.0\%$ for RoadGen.
RoadWeaver produces road networks with more closed-loop structures, resulting in more connected road blocks and multiple routing possibilities.

RoadWeaver also achieves the smallest endpoint alignment error, with only $0.24$ m, while the baselines range from $4.32$ m to $8.17$ m.
The smaller error indicates that lane endpoints remain better aligned during graph expansion, resulting in more consistent road connections.

These results show that RoadWeaver can generate large-scale HD maps from scratch while maintaining graph connectivity and lane-level geometric consistency.

\subsection{Scalability and Efficiency}

We compare the computational efficiency of all methods with respect to the number of generated road-network nodes, which provides a unified measure since different methods define map scale differently.

The scalability of the baselines is constrained by their generation mechanisms. 
HDMapGen does not impose an explicit limit on scale, but as is shown in Figure~\ref{fig:quality}, larger graphs provide limited quality improvement.
RoadGen is limited by geometric feasibility.
Once the available connection queue is exhausted, the assembly process terminates early, so valid maps rarely extend beyond roughly 60 nodes.
MetaDrive scales by increasing the predefined map configuration, which leads to rapidly rising resource consumption as the map grows.
In contrast, RoadWeaver scales by enlarging the map area while maintaining a bounded road density, so the number of generated nodes can grow naturally without changing the generation procedure or relying on a fixed map template.

\begin{figure}[htb]
    \centering
    \includegraphics[width=\linewidth]{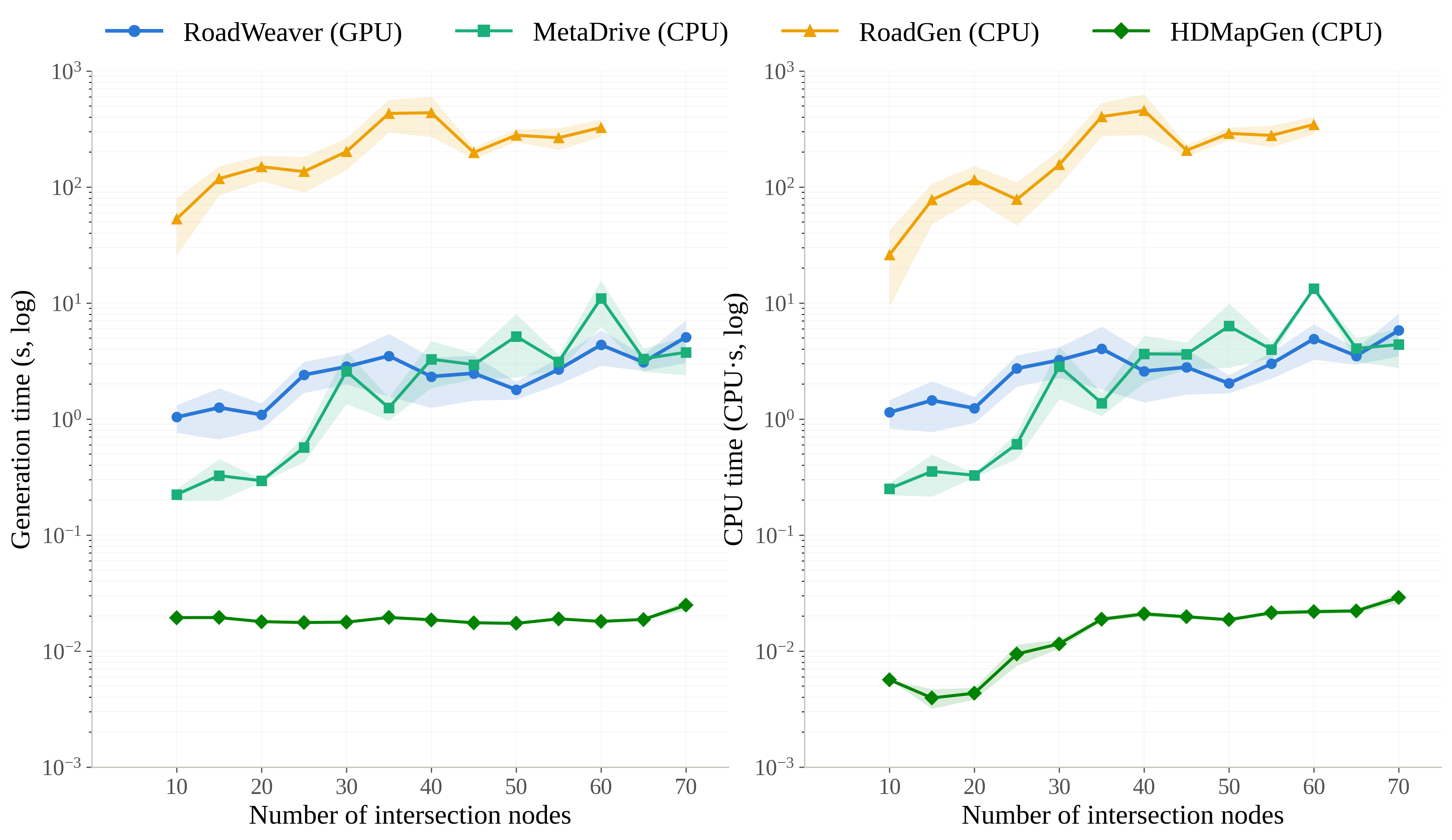}
    \caption{Generation time and CPU consumption of different methods as the number of graph nodes increases. CPU consumption is measured in CPU-seconds, calculated as the product of the average CPU utilization and the wall-clock generation time.}\label{fig:efficiency}
\end{figure}

Figure~\ref{fig:efficiency} shows the generation time and CPU consumption as the number of graph nodes increases.
HDMapGen can increase graph size, but larger graphs provide limited quality improvement.
RoadGen maintains strong geometric constraints, but its generation time increases from approximately 50 s to 370 s, making large-scale expansion impractical.
Compared with MetaDrive, RoadWeaver achieves comparable runtime while offering better scalability and road-network quality.
Specifically, RoadWeaver generates maps within approximately $1.39$--$3.50$~s.

\subsection{Simulator Deployment Validation}

To validate the practical usability of the generated HD maps, we integrate RoadWeaver into Tactics2D~\cite{li2024tactics2d}. 
The generated maps are compatible with OSM and OpenDRIVE representations and can be imported into Tactics2D without manual conversion.
Tactics2D is a lightweight open-source simulator that supports CPU-based closed-loop autonomous driving evaluation and provides lane-level HD map representation, route planning, vehicle models, driving behaviors, and benchmark datasets.
These characteristics make it an appropriate platform for verifying whether RoadWeaver can be directly deployed for autonomous driving simulation.

\begin{figure}[htb]
    \centering
    \subfloat[]{\includegraphics[width=0.47\linewidth]{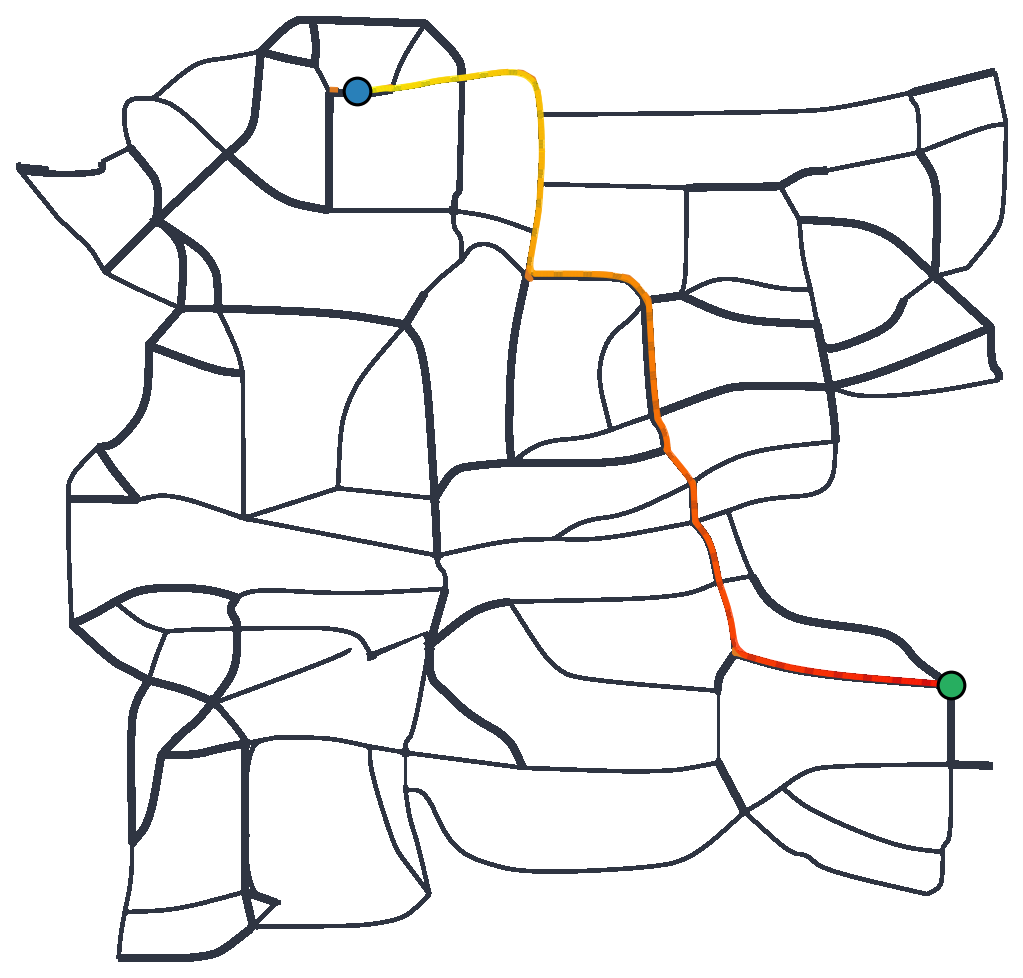}}
    \hfill
    \subfloat[]{\includegraphics[width=0.47\linewidth]{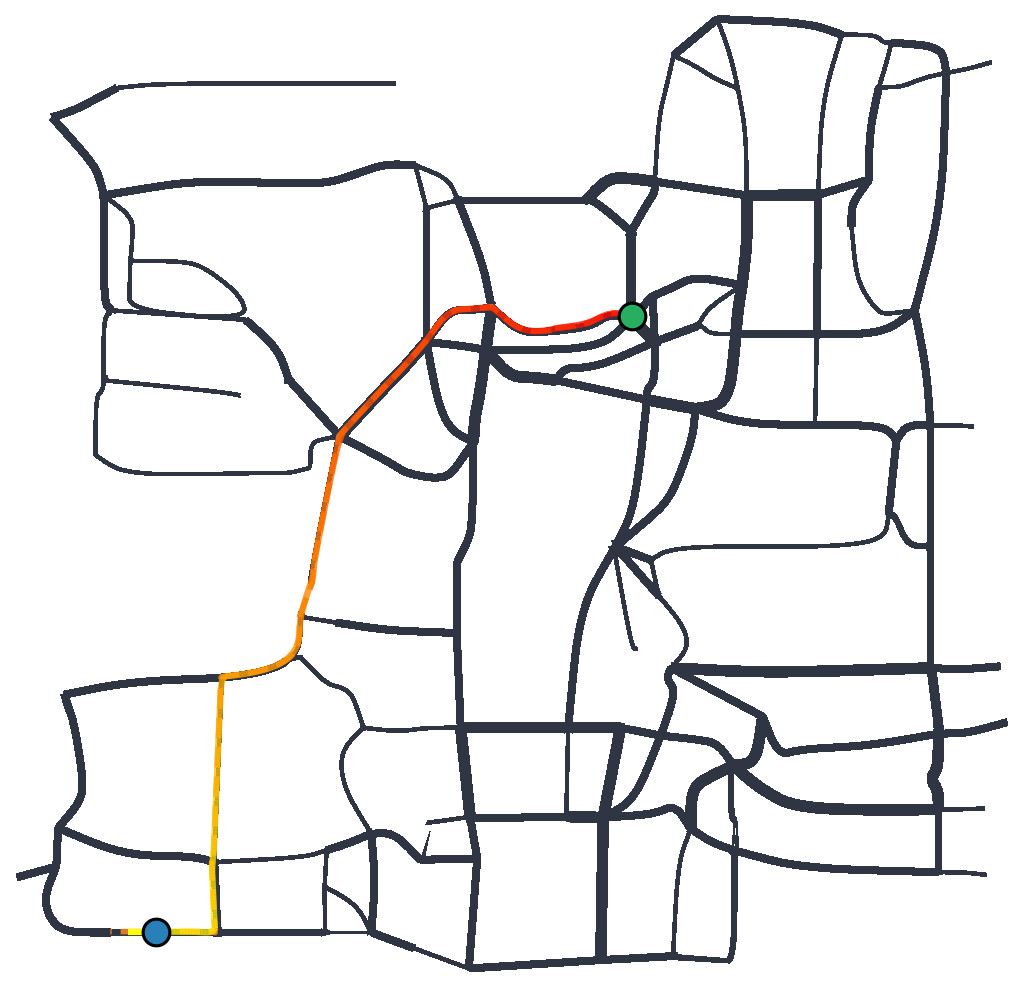}}
    \caption{Routing sample of generated maps from RoadWeaver in Tactics2D.}\label{figure:routing}
\end{figure}

We randomly generate 100 lane-level HD maps covering diverse road layouts and network densities.
Each generated map contains complete lane-level topology, including predecessor, successor, and neighboring lane relationships, and is directly imported into Tactics2D without any manual editing or post-processing.
All generated maps are successfully imported into the simulator, achieving an import success rate of 100.0\%.
We further perform 10 randomly sampled routing tasks on each map, resulting in a total of 1,000 route-planning tasks.
The built-in route planner successfully generates valid paths for most of the tasks, corresponding to a routing success rate of 98.7\%.
The average routing time is about $0.82 \pm 0.17$~s.
Representative generated maps and their corresponding routing results are shown in Fig.~\ref{figure:routing}.

These results demonstrate that the generated HD maps are not only geometrically and topologically valid, but also directly deployable for downstream autonomous driving simulation and route-planning tasks without manual intervention.

\section{Conclusion}\label{section:conclusion}

This paper presented RoadWeaver, a coarse-to-fine framework for generating large-scale lane-level HD maps from scratch.
Unlike existing methods, RoadWeaver generates simulation-ready maps while preserving road topology and lane-level geometry.
Experiments show that the generated maps are diverse, scalable, and directly compatible with existing driving simulators, making them suitable for large-scale closed-loop evaluation of ADS.
Future work will focus on richer traffic semantics, roadside assets, and interactive map editing, and further improve the realism of generated driving environments.

\bibliographystyle{IEEEtran}
\bibliography{main}

\end{document}